\documentclass[10pt,journal,compsoc]{IEEEtran}
\usepackage[latin9]{inputenc}
\usepackage{color}
\usepackage{array}
\usepackage{verbatim}
\usepackage{units}
\usepackage{multirow}
\usepackage{amsmath}
\usepackage{graphicx}
\usepackage{ragged2e}
\usepackage{amssymb}

\makeatletter

\DeclareTextSymbolDefault{\textquotedbl}{T1}
\providecommand{\tabularnewline}{\\}

\let\oldforeign@language\foreign@language
\DeclareRobustCommand{\foreign@language}[1]{%
  \lowercase{\oldforeign@language{#1}}}

\ifCLASSOPTIONcompsoc
  \usepackage[nocompress]{cite}
\else
  \usepackage{cite}
\fi
\ifCLASSINFOpdf
\else
\fi

\makeatother

\begin{document}
\title{Dataset Bias in Few-shot Image Recognition}
\author{Shuqiang~Jiang,~\IEEEmembership{Senior~Member,~IEEE},~Yaohui Zhu,~Chenlong
Liu,\protect\\
 Xinhang Song,~Xiangyang Li,~and Weiqing Min \IEEEcompsocitemizethanks{\IEEEcompsocthanksitem
The authors are with the Key Laboratory of Intelligent Information
Processing of Chinese Academy of Sciences (CAS), Institute of Computing
Technology, CAS, Beijing 100190, China and also with the University
of Chinese Academy of Sciences, Beijing 100049, China.\protect\\
 E-mail: sqjiang@ict.ac.cn, \{yaohui.zhu, chenlong.liu, xinhang.song,
xiangyang.li\}@vipl.ict.ac.cn, minweiqing@ict.ac.cn.}}

%

\markboth{Journal of \LaTeX\ Class Files,~Vol.~14, No.~8, August~2015}{Shell \MakeLowercase{\textit{et al.}}: Bare Demo of IEEEtran.cls
for Computer Society Journals}

\IEEEtitleabstractindextext{
\begin{abstract}
\justifying The goal of few-shot image recognition (FSIR) is to identify
novel categories with a small number of annotated samples by exploiting
transferable knowledge from training data (base categories). Most
current studies assume that the transferable knowledge can be well
used to identify novel categories. However, such transferable capability
may be impacted by the dataset bias, and this problem has rarely been
investigated before. Besides, most of few-shot learning methods are
biased to different datasets, which is also an important issue that
needs to be investigated deeply. In this paper, we first investigate
the impact of transferable capabilities learned from base categories.
Specifically, we use the relevance to measure relationships between
base categories and novel categories. Distributions of base categories
are depicted via the instance density and category diversity. The
FSIR model learns better transferable knowledge from relevant training
data. In the relevant data, dense instances or diverse categories
can further enrich the learned knowledge. Experimental results on
different sub-datasets of ImageNet demonstrate category relevance,
instance density and category diversity can depict transferable bias
from distributions of base categories. Second, we investigate performance
differences on different datasets from the aspects of dataset structures
and different few-shot learning methods. Specifically, we introduce
image complexity, intra-concept visual consistency, and inter-concept
visual similarity to quantify characteristics of dataset structures.
We use these quantitative characteristics and eight few-shot learning
methods to analyze performance differences on five different datasets
(i.e., MiniCharacter, MiniImageNet, MiniPlaces, MiniFlower, MiniFood).
Based on the experimental analysis, some insightful observations are
obtained from the perspective of both dataset structures and few-shot
learning methods. We hope these observations are useful to guide future
few-shot learning research on new datasets or tasks. Our data is available
at http://123.57.42.89/dataset-bias/dataset-bias.html. 
\end{abstract}

}
\maketitle

\IEEEraisesectionheading{\section{Introduction}\label{sec:introduction}}

%
%
%
%
\IEEEPARstart{L}{earning} from few examples and generalizing to different
situations are capabilities of human visual intelligence. During the
past years, significant progress has been made on image recognition
\cite{he2016deep,szegedy2017inception,huang2017densely} with the
assistance of deep learning techniques \cite{lecun2015deep} and large
scale labelled dataset \cite{deng2009imagenet,zhou2018places}. However,
this kind of human visual intelligence is yet to be matched by leading
machine learning models. Humans can easily learn to recognize a novel
object category after glancing at only one or a few examples \cite{thrun1998learning}.
This cognitive ability can be explained by the \textquotedbl learning
to learn\textquotedbl{} mechanism in the brain \cite{harlow1949formation},
which means that human can make inference so that their previously
acquired knowledge on related tasks can be flexibly adapted to a new
task. Inspired by this human ability, the few-shot image recognition
(FSIR) is proposed to learn novel concepts from a few, or even a single
example.

The task of FSIR establishes a new recognition setup to transfer the
knowledge of training tasks sampled from training (base) categories
to the new task with one or very few samples. Instead of learning
one single recognition task, most FSIR models learn plenty of recognition
tasks. Each task contains a support set (training samples) and a target
set (test samples). The support set consists of a few available labelled
data, which is exploited to learn a task-specific model. Then the
learned model is evaluated on the target set. Each task in these two
sets shares the same concepts. But concepts of testing tasks come
from novel categories, which are different from those of training
tasks.

Current studies of FSIR \cite{snell2017prototypical,sung2018learning,chen2019closer,lee2019meta,Finn2017Model,Munkhdalai2018Rapid,T-Fine-2020}
achieve transferable knowledge by learning training tasks or base
categories. These studies mostly focus on transferable knowledge between
datasets or tasks by exploiting given base categories. The majority
of current works assume the transferable knowledge globally shared
across all tasks, and consider that the leaned knowledge can be well
adapted to novel categories. However, transferable knowledge is highly
dependent on the distributions of base categories. FSIR models can
acquire biased transferable capabilities if distributions of base
categories and novel categories are very different. Furthermore, current
studies rarely explore the characteristics of dataset structures,
which include not only an amount of information in the image but also
semantic gaps between original images and concepts. Current works
do not deeply dig differences in dataset structures. Therefore, current
few-shot learning methods may be biased to different datasets. 
\begin{figure*}
\begin{centering}
\includegraphics[scale=0.47]{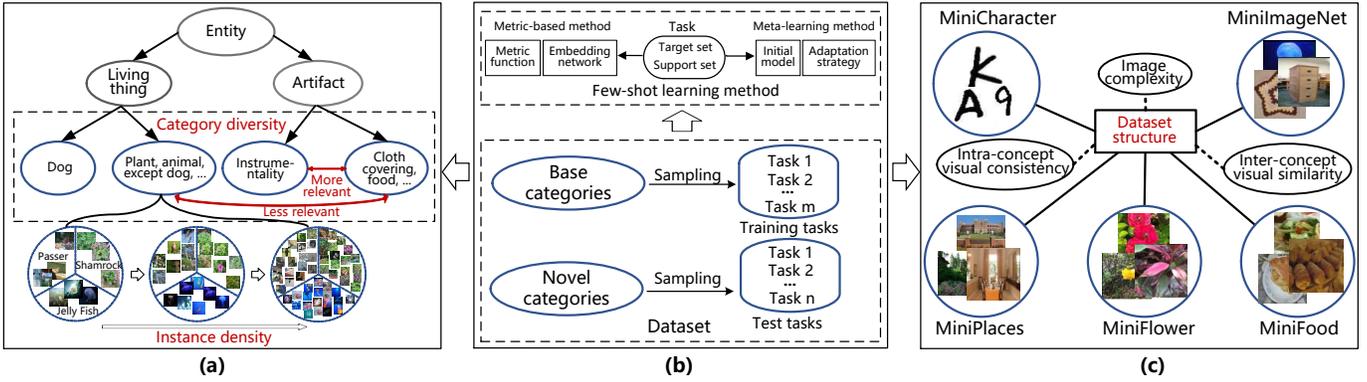} 
\par\end{centering}
\caption{\label{fig:The-two-investigations}The two investigations for FSIR
from the dataset. (a): illustrations of dataset diversity. (b): illustrations
of the few-shot learning tasks sampled from datasets and the few-shot
learning method. (c): illustrations of dataset structures.}
\end{figure*}

Two problems arise based on the above analysis: i) What factors can
describe transferable bias from distributions of base categories?
ii) What characteristics can depict bias of few-shot learning methods
on different datasests? In this paper, we focus on studying these
two problems systematically, which have rarely been explored before.
Fig. \ref{fig:The-two-investigations} (a) illustrates the investigation
of the first problem, and Fig. \ref{fig:The-two-investigations} (c)
illustrates the investigation of the second problem.

For the first problem, we aim to obtain transferable knowledge from
base categories, which can be better adapted to novel categories.
The FSIR model can learn more accurate knowledge from dense instances,
more comprehensive knowledge from diverse categories, especially,
transferable capabilities from relative categories. Therefore, we
introduce the dataset diversity to depict distributions of base categories
and the relevance to measure relationships between base categories
and novel categories. The dataset diversity contains instance density
and category diversity. We conduct experiments on eight few-shot learning
methods, which contains four classic methods such as Prototypical
Net {[}9{]}, MAML {[}13{]}, and four recent methods. We measure relevance
of categories contained in ImageNet \cite{russakovsky2015imagenet}
both qualitatively and quantitatively to obtain different sub-datasets
which contain dense instances and diverse categories. Under the settings
of different relevance, instance density and category diversity are
explored respectively. Besides, we further compare instance density
and category diversity with the fixed number of total samples.

For the second problem, we aim to analyze differences in performance
on different datasets from characteristics of dataset structures and
different few-shot learning methods (e.g., metric-based methods and
meta-learning methods shown in Fig. \ref{fig:The-two-investigations}
(b)). We introduce image complexity, intra-concept visual consistency,
and inter-concept visual similarity to quantify characteristics of
dataset structures. To conduct comprehensive evaluations on multiple
datasets, we introduce five datasets, including simple character images
(i.e., MiniCharacter), images with different number of objects (i.e.,
MiniImagenet, MiniPlaces), and two fine-grained datasets (i.e., MiniFlower
and MiniFood). We use five kinds of features to calculate intra-concept
visual consistency and inter-concept visual similarity, and measure
image complexity in two manners. These quantitative characteristics
are used to analyze differences in performance on different datasets.
In addition, we give analysis on differences in performance of these
few-shot learning methods.

In summary, our main contributions are as follows: i) We systematically
investigate the influence of knowledge learned from base categories.
ii) We systematically investigate differences in performance on different
datasets with three characteristics of dataset structures and two
types of few-shot learning methods. iii) Based on the above investigates,
we can obtain following key conclusions: 
\begin{itemize}
\item The FSIR model can obtain better performance with knowledge learned
on relevant base categories rather than irrelevant ones.
\item The FSIR model can obtain improvement with knowledge learned on more
dense instances or diverse categories without reducing the relevance.
\item The FSIR model can obtain more improvements with knowledge learned
on diverse categories than that learned on dense instances without
reducing the relevance, when there are enough instances for each category.
\item The FSIR model obtains different performance on different datasets,
which can be depicted with image complexity, intra-concept visual
consistency, and inter-concept visual similarity.  
\item Different FSIR models have diverse performance on different datasets,
which is relevant to both dataset structures and the method ability.
The method ability is closely related to characteristics of dataset
structures. 
\end{itemize}
The remainder of this paper is constructed as follows. Sect. \ref{sec:Related-Work}
provides the related work, including FSIR, domain adaptation, and
few-shot domain adaptation. Sect. \ref{sec:Preliminary} gives formulation
of FSIR and reviews two types of classic few-shot learning methods.
Sect. \ref{sec:Evaluation-of-Diversity} presents evaluations of the
dataset diversity in detail. Sect. \ref{sec:Evaluation-of-Dataset}
presents evaluations of the dataset structure and experimental analysis
in detail. Sect. \ref{sec:Perspectives-and-future} provides some
perspectives and future directions. Finally, the paper closes with
conclusions in Sect. \ref{sec:Conclusions}.

\section{\label{sec:Related-Work}Related Work}

\subsection{Few-shot Image Recognition}

The goal of FSIR is to identify novel categories with a few annotated
examples and knowledge obtained from base categories. In early attempts,
Fei-Fei $et\,al.$ \cite{fei2006one}\ propose a variational bayesian
framework for one-shot image classification, and Lake $et\,al.$ \cite{Lake2015Human}
propose hierarchical bayesian program learning on the few-shot alphabet
recognition tasks. Inspired by architectures with augmented memory
capacities such as Neural Turing Machines (NTMs), Santoro $et\,al.$
\cite{santoro2016meta} propose a meta-learning method with memory-augmented
neural networks. Afterwards, there are three kinds of methods to deal
with the FSIR problem. The first one is metric-based method (i.e.,
learning to compare), which learns a transferable embedding network
or function. This function transforms the original images into the
embedding space such that these images can be recognized with a nearest
neighbor \cite{koch2015siamese,zhang2020deepemd}, a linear classifier
\cite{vinyals2016matching,snell2017prototypical,simon2020adaptive}
or a deep nonlinear metric \cite{sung2018learning}. The second one
is meta-learning method \cite{schmidhuber1987evolutionary,naik1992meta,lee2019meta}
(i.e., learning to learn), whose learning occurs at two levels: task-level
learning and take-specific adaption. The task-level learning is usually
implemented by an additional meta-learner \cite{Munkhdalai2017Meta,Munkhdalai2018Rapid}
or a sensitive initialization shared with task-specific learners \cite{Finn2017Model,meta-2020},
which can provide meta-level information for the take-specific adaption.
The third method is generative or augmentation-based method (i.e.,
learning to generate or augment), which learns a generative model
to synthesize more samples and then trains a robust classifier. This
generative model uses semantic information \cite{Dixit2017AGA,chen2018semantic}
(e.g., attribute), or base categories for analogy or hallucination
\cite{hariharan2017low,schwartz2018delta}.

Recently, some works study FSIR from the view of self-supervised approaches
\cite{gidaris2019boosting,li2019learning} and semi-supervised approaches
\cite{ren2018meta,liu2019learning}. Yu\ $et\,al.$\ \cite{yu2020transmatch}\ propose
a two-stage approach which explores knowledge from both annotated
examples of base categories and un-annotated ones of novel categories.
The above works focus on learning transferable knowledge with given
datasets. However, we investigate the performance of FSIR from dataset
diversity with changeable base categories and different characteristics
of dataset structures. A more related work is \cite{sbai2020impact},
which shows that increasing relevant categories in close or far semantic
distances can boost the performance of FSIR. In addition, our work
also considers increasing irrelevant categories, and experimental
results illustrate that more irrelevant categories cannot improve
the performance, suggesting that it's not the more categories the
better performance. Furthermore, we investigate differences in performance
on different datasets from the dataset structure and different few-shot
learning methods, which is not explored by \cite{sbai2020impact}.

\subsection{Domain Adaptation}

Domain adaptation utilizes labeled data in one or more relevant source
domains to execute new tasks in a target domain with scarce annotated
data. It aims to solve the domain gap and transfer knowledge learned
on the source domain to the target domain\ \cite{AD-2018,AD-2017,AD-2015-backpropagation,DA_2015}.
As many approaches are based on deep neural networks, Li $et\,al.$
\cite{li2019multifaceted} construct source and target datasets with
various distances to analyze factors that affect the effectiveness
of using prior knowledge from a pre-trained model with a fixed network
architecture. Azizpour $et\,al.$ \cite{AA-2016} investigate factors
(e.g., network architectures, parameters of feature extraction) affecting
the transferability of feature representations in generic convolutional
networks. To learn domain invariant features Minimizing the domain
discrepancy, Long $et\,al.$ \cite{long2016unsupervised} propose
a deep network architecture that can jointly learn adaptive classifiers
and transferable features from labeled data in the source domain and
unlabeled data in the target domain. Meanwhile, with significant advances
made in image synthesis by generative adversarial networks, many methods
focus on learning domain-independent features and synthesizing examples
in the new domain\cite{Tzeng_2017_CVPR,zhu2017unpaired}. Hoffman
$et\,al.$ \cite{AD-2018-GAN}\ propose adversarial learning method
that utilizes both generative image space alignment and latent representation
space alignment. Zhang $et\,al.$ \cite{AD-CVPR2019} propose an adversarial
learning method with two-level domain confusion losses. Cui $et\,al.$\ \cite{cui2020gradually}
propose gradually vanishing bridge mechanism to learn more domain-invariant
representations. To tackle predictive domain adaptation, Mancini $et\,al.$
\cite{AD-Gaph-2019} leverage meta data information to build a graph
and design novel domain-alignment layers based on the graph for domain
adaption. These works have the same classes among different domains.
However, we address the problem in FSIR, where the classes in target
domain are disjoint with ones in source domains. Meanwhile, the training
examples in the target domain are limited or rare.

\subsection{Few-shot Domain Adaptation}

Few-shot domain adaptation aims to recognize novel categories with
a few annotated data in the target domain and sufficient data in the
source domains. Some works \cite{motiian2017few-shot,kang2018transferable}
assume that the target domain contains the same classes as the source
domain. Recently, some efforts attempt to address a more flexible
and challenging few-shot domain adaptation, where the target domain
and source domains have disjoint classes, called cross-domain few-shot
learning. Chen $et\,al.$ \cite{chen2019closer} evaluate current
models and proposed baselines on cross-domain few-shot protocols (from
MiniImagnet \cite{vinyals2016matching} to cub \cite{welinder2010caltech}).
Tseng $et\,al.$ \cite{tseng2020cross} propose a learned feature-wise
transformation to stimulate feature distributions cross domains with
a small number of samples. Guo $et\,al.$\cite{new-bench-2019} introduce
a more realistic cross-domain few-shot learning, where the source
domain consists of common images from Imagenet \cite{russakovsky2015imagenet},
and the target domains contain rare images such as satellite images
and radiological images. Besides, Vuorio $et\,al.$ \cite{vuorio2019multimodal}
propose the multi-domain few-shot learning, and use a task-aware modulation
to promote the learning of meta-learner. Yao $et\,al.$ \cite{yao2019hierarchically}
propose a hierarchically structured meta-learning algorithm to promote
knowledge customization on different domains but simultaneously preserve
knowledge generalization in related domains. Triantafillou $et\,al.$
\cite{meta-2020} introduce a meta-dataset that consists of 10 datasets
in different domains and present experimental evaluation of current
models and proposed baselines. These works only use given source domains
without selections, in contrast, we selectively use source domains
(i.e., base categories) and systematically investigate different target
domains from characteristics of dataset structures and different few-shot
learning methods.

\section{\label{sec:Preliminary}Preliminaries}

\subsection{Few-shot Image Recognition Formulation}

In the regular machine learning setting, a classification problem
is defined on instances $\mathcal{I}_{(x,y)}\sim p(\mathcal{I})$,
where $x$ is a sample and $y$ is the corresponding label. But most
FSIR models learn task instances $\mathcal{T}\sim p(\mathcal{T})$.
Sampling from training and test data, we form training tasks $\mathcal{T^{\mathit{train}}}$
and testing tasks $\mathcal{T^{\mathit{test}}}$, respectively. According
to existing settings, the training and test data are made of base
and novel categories, respectively, and each category has plenty of
samples. For example, in MiniImageNet \cite{vinyals2016matching},
the number of base and novel categories are 64 and 20 respectively,
where each category has 600 samples.

\begin{figure}
\noindent \begin{centering}
\includegraphics[scale=0.305]{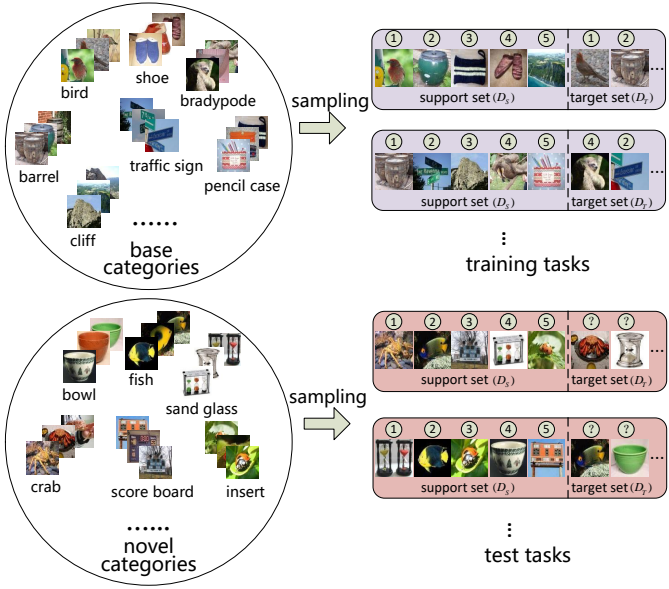} 
\par\end{centering}
\caption{\label{fig:The-training-and test tasks}The training and test tasks
are formed by randomly sampling from the base and novel categories,
respectively.}
\end{figure}

In training tasks $\mathcal{T^{\mathit{train}}}$ or testing tasks
$\mathcal{T^{\mathit{test}}}$, each task is defined as $\mathcal{T_{\mathit{j}}}=\{D_{\mathcal{T}_{j},S},D_{\mathcal{T}_{j},T}\}$,
where $D_{\mathcal{T}_{j},S}$ is a support set (training samples)
and $D_{\mathcal{T}_{j},T}$ is a target set (test samples). The support
set $D_{\mathcal{T}_{j},S}=\{(x_{c}^{k}\text{,}y_{c}^{k})\mid c=1,2,...,C;k=1,2,...,K\}$
and the target set $D_{\mathcal{T}_{j},T}$ consist of $C$ categories
randomly sampled from the total categories, and each sampled category
contains $K$ labeled samples in the support and some samples in the
target set. And this task is called $C$-way $K$-shot task. Test
tasks $\mathcal{T^{\mathit{test}}}$ and training tasks $\mathcal{T^{\mathit{train}}}$
have the same form but with disjoint label space since they have different
categories. Fig. \ref{fig:The-training-and test tasks} illustrates
the 5-way 1-shot training and test tasks, which are sampled from base
and novel categories, respectively.

\subsection{Few-shot Learning Methods}

Metric-based methods and meta-learning methods are employed to explore
the dataset bias. These two kinds of methods do not use additional
information and are easy to be implemented with less options, compared
with generative or augmentation-based methods. In the following, we
introduce the typical works in these two kinds of methods.

\subsubsection{Metric-based Methods}

This kind of method contains two parts: an embedding network or function
$\mathcal{G}()$ and a metric function $\mathcal{M}()$. A key assumption
is that $\mathcal{G}()$ learns domain-general information as an inductive
bias \cite{thrun2012learning} to generalize novel categories. In
addition, the learning target or the loss function $\mathcal{L}()$
affects $\mathcal{G}()$ learning (or $\mathcal{M}()$ learning, when
$\mathcal{M}()$ is parameterized). Hence, we review the following
metric-based few-shot leaning methods from $\mathcal{L}()$ and $\mathcal{M}()$.

\textbf{Prototypical Net (PN)} \cite{snell2017prototypical}. In this
method, the $\mathcal{M}()$ is the Euclidean distance function. The
$\mathcal{L}()$ is the cross entropy loss, and the probability of
each sample in $D_{T}$ is defined as (omitting the index of task):
\begin{equation}
P(\overline{y}|\overline{x},D_{S})=\frac{e^{-\mathcal{M}(\mathcal{G}(\overline{x}),\underset{(x,y)\in D_{S}}{\sum}\frac{\mathbf{I\{}y=\overline{y}\}\mathcal{G}(x)}{K})}}{\sum_{a=1}^{a=C}e^{-\mathcal{M}(\mathcal{G}(\overline{x}),\underset{(x,y)\in D_{S}}{\sum}\frac{\mathbf{I\{}y=y{}_{a}\}\mathcal{G}(x)}{K})}}
\end{equation}
where $C$ is the number of way, $K$ is the number of shot and $(\overline{x},\overline{y})\in D_{T}$.

\textbf{Relation Network (RN)} \cite{sung2018learning}. This method
learns a parameterized $\mathcal{M}()$, which is implemented by a
neural network. The $\mathcal{L}()$ is mean square error, defined
as: 
\begin{equation}
\mathcal{L}()=\sum_{(\overline{x},\overline{y})\in D_{T}}\sum_{(x,y)\in D_{s}}(r(\overline{y}|\overline{x},D_{S})-\mathbf{I\{}y=\overline{y}\})^{2}
\end{equation}
where $r(\overline{y}|\overline{x},D_{S})=\mathcal{M}(\mathcal{G}(\overline{x}),pooling(\{\mathcal{G}(x)\mid(x,y)\in D_{s},y=\overline{y}\})$,
and the $pooling()$ is maxpooling in \cite{sung2018learning}.

\textbf{Deep Subspace Networks (DSN)} \cite{simon2020adaptive}. In
this method, the $\mathcal{M}()$ is implemented with task-adaptive
subspace metrics, i.e., $\{\mathcal{M_{\mathit{y_{a}}}}(\overline{x})\}_{a=1}^{a=C}$.
The $\mathcal{L}()$ is the cross entropy loss, and the probability
of each sample in $D_{T}$ is defined as: 
\begin{equation}
P(\overline{y}|\overline{x},D_{S})=\frac{e^{-\mathcal{M_{\mathit{\overline{y}}}}(\overline{x})}}{\sum_{a=1}^{a=C}e^{-\mathcal{M_{\mathit{y_{a}}}}(\overline{x})}}\label{eq:dsn}
\end{equation}
where $\mathcal{M_{\mathit{y_{a}}}}(\overline{x})=-\parallel(\mathrm{I}-\mathrm{\mathit{P_{y_{a}}P_{y_{a}}^{\mathrm{T}}}})(\mathcal{G}(x)-u_{y_{a}})\parallel^{2}$,
$P_{y_{a}}$ is a matrix with orthogonal basis for the linear subspace
spanning $X_{a}=\{\mathcal{G}(x)\mid y=y_{a},(x,y)\in D_{s}\}$, $u_{y_{a}}=\frac{1}{K}\sum_{x\in X_{a}}\mathcal{G}(x)$,
and $(\overline{x},\overline{y})\in D_{T}$

\textbf{DEMD} \cite{zhang2020deepemd}. In this method, the Earth
Mover's Distance (EMD) is to obtain optimal matching flows $\mathcal{\tilde{M}}()$
between each two feature maps ($\mathcal{G}()$). Thus the similarity
metric between two maps is $\mathcal{M}()=Cos()\mathcal{\tilde{M}}()$,
where $Cos()$ is cosine similarity of two vectors in the maps. The
$\mathcal{L}()$ is the cross entropy loss, and the probability of
each sample in $D_{T}$ is defined as:

\begin{equation}
P(\overline{y}|\overline{x},D_{S})=\frac{e^{-\mathcal{M}(\mathcal{G}(\overline{x}),\underset{(x,y)\in D_{S}}{\sum}\mathbf{I\{}y=\overline{y}\}\mathcal{G}(x))}}{\sum_{a=1}^{a=C}e^{-\mathcal{M}(\mathcal{G}(\overline{x}),\underset{(x,y)\in D_{S}}{\sum}\mathbf{I\{}y=y{}_{a}\}\mathcal{G}(x))}}
\end{equation}
where $(\overline{x},\overline{y})\in D_{T}$.

\subsubsection{Meta-learning Methods}

This kind of method usually contains two parts: an initial model $\mathcal{F}(;\theta)$
and an adaptation strategy $\mathcal{S}(;\delta)$, where $\theta$
are parameters of $\mathcal{F}()$, and $\delta$ are parameters of
$\mathcal{S}()$. The processes of this kind of method are: i) computing
the gradient (or loss) information on support set: $grad_{a_{t}/\theta}=\nabla_{a_{t}/\theta}\mathcal{L}(\mathcal{F}(D_{S}(x);\theta),D_{S}(y))$,
where $\mathcal{L}(,)$ is the loss function, $a_{t}$ is the $t^{th}$
neurons of $\mathcal{F}()$; ii) transforming the gradient information
into adaptive information: $I=S(grad;\delta)$; iii) leveraging the
adaptive information to obtain an updated model: $\mathcal{B\mathrm{(}\mathcal{F}\mathrm{(}\mathrm{;}\theta\mathrm{)}\mathrm{,}\mathrm{\mathit{I}}\mathrm{)}}$.
Similarly, we review the following meta-leaning methods from the $\mathcal{S}()$.
Generally speaking, the final learning target or the loss function
calculated on the target set has the same formulation as $\mathcal{L}(,)$
mentioned in the ii) process.

\textbf{Model-Agnostic Meta-Learning} (\textbf{MAML}) \cite{Finn2017Model}.
This method is inspired by fine-tuning. It computes gradient of the
whole parameter $\theta$, and then directly uses the gradient on
original parameters with one or a few gradient steps to obtain the
updated model: $\mathcal{B\mathrm{(}\mathcal{F}\mathrm{(}\mathrm{;}\theta\mathrm{)}\mathrm{,}\mathrm{\mathit{I}}\mathrm{)}}\mathcal{=F}\mathrm{(}\mathrm{;}\theta-r\cdot I\mathrm{)}$,
where $I=grad_{\theta}$, $r$ is updating learning rate.

\textbf{Proto-MAML} \cite{meta-2020}. As a variant of MAML, this
method also updates the whole parameter $\theta$ for adaptation.
Different from MAML, the classifier weight of Proto-MAML is initialized
with the prototype of classes \cite{snell2017prototypical}. The classifier
weight initialization can be formed as $\theta_{c}:=\mathcal{F}(D_{S}(x);\theta_{e})$,
where $\theta_{c}$ and $\theta_{e}$ are parameters of the classifier
and the embedding network, respectively. The gradient on support set
is achieved via $grad_{\theta}=\nabla_{\theta}\mathcal{L}(\mathcal{F}(D_{S}(x);\{\theta_{e},\theta_{c}\}),D_{S}(y))$,
where $\theta=\{\theta_{e},\theta_{c}\}$. Thus the updated model
is $\mathcal{B\mathrm{(}\mathcal{F}\mathrm{(}\mathrm{;}\theta\mathrm{)}\mathrm{,}\mathrm{\mathit{I}}\mathrm{)}}\mathcal{=F}\mathrm{(}\mathrm{;}\theta-r\cdot grad_{\theta}\mathrm{)}$,
where $r$ is updating learning rate.

\textbf{adaCNN} \cite{Munkhdalai2018Rapid}. This method computes
gradient of neurons $grad_{t}=\nabla_{a_{t}}\mathcal{L}(\mathcal{F}(D_{S}(x);\theta),D_{S}(y))$,
and transforms the gradient into conditional shift vectors $\beta_{t,m}=I_{m,t}=S(grad_{a_{t,m}};\delta)$
($m$ is the $m^{th}$ instance in the support set) that are saved
in a memory. The updated model is as follows: 
\begin{equation}
\mathcal{B\mathrm{(}\mathcal{F}\mathrm{(}\mathrm{;}\theta\mathrm{)}\mathrm{,}\beta_{\mathit{t}}})=\begin{cases}
\mathcal{\sigma\mathrm{(}F}\mathrm{(\mathit{a}_{\mathit{t}}}\mathrm{;}\theta\mathrm{)})+\sigma(\beta_{t}) & t\neq\mathrm{T}\\
softmax(\mathcal{F}\mathrm{(\mathit{a}_{\mathit{t}}}\mathrm{;}\theta\mathrm{)}+\beta_{t}) & t=\mathrm{T}
\end{cases}
\end{equation}
where $\mathrm{T}$ is the final layer, $\sigma()$ is a nonlinear
function, and $softmax()$ is the Softmax function. The layer-wise
shifts are recalled from memory via a soft attention to obtain $\beta_{t}$
($\beta_{t}=\underset{m}{\sum}w_{m}\beta_{t,m}$, $w_{m}$ is calculated
by a key function), which is used for adjusting the output of initial
model.

\textbf{MetaOpt }\cite{lee2019meta}. This method learns an adaptive
classifier for adaptation. The classifier is implemented with SVM,
which can be solved by optimizing a convex objective function. The
weight of the adaptive classifier is achieved via $\theta_{c}^{*}:=\underset{\theta_{c}}{argmin}\mathcal{L}(\mathcal{F}(D_{S}(x);\{\theta_{e},\theta_{c}\}),D_{S}(y))$,
where $\theta_{c}^{*}$ and $\theta_{e}$ are parameters of the adaptive
classifier and the embedding network, respectively. Thus the updated
model is $\mathcal{B\mathrm{(}\mathcal{F}\mathrm{(}\mathrm{;}\theta\mathrm{)}\mathrm{,}\mathrm{\mathit{I}}\mathrm{)}}\mathcal{=F}\mathrm{(}\mathrm{;}\{\theta_{e},\theta_{c}^{*}\mathrm{))}$.

\subsubsection{Summary}

Metric-based methods do not need task-specific adaptation. They require
data with a high relevance between base categories and novel categories
especially for a learnable metric. Metric-based methods focus on designing
an effective metric. For the above four metric-based methods, two
of them are classic, anther two are up to date. In the two classic
method, PN uses a predefined metric, and RN utilizes a parametric
network as the metric. In the two recent works, DSN adopts task-adaptive
subspace metrics, and DEMD measures each two images with the Earth
Mover's Distance. Therefore, the four metric-based methods have their
characteristics in the metric function.

Compared with metric-based methods, meta-learning methods update their
models via task-specific adaptation, such that these methods are less
dependent on the relevance. For the above four meta-learning methods,
MAML is the classic one, which updates the whole weights of the initial
network for adaption. Different from the MAML, adaCNN updates activations
for adaption. Proto-MAML is a variant of MAML, and initializes classifier
weight with the prototype of classes. For adaption, the three meta-learning
methods update the whole weight or some activations of network. In
contrast, MetaOpt only updates task-sensitive classifier. Therefore,
the four meta-learning methods have their characteristics in the model
initialization or the designing of adaptation strategy.

 We use the eight representative few-shot learning methods (i.e.,
four metric-based methods and four meta-learning methods) to carry
out experiments on diverse datasets to analyze dataset bias of FSIR
in the following sections.

\section{\label{sec:Evaluation-of-Diversity}Evaluation of Dataset Diversity}

The FSIR model aims to recognize novel categories with a small amount
of samples by exploiting learnable generic knowledge. This kind of
knowledge is learned from sufficient base categories, whose diversity
can explicitly affect the quality of the learned knowledge. In this
section, we study dataset diversity of base categories to explore
FSIR. First, we introduce some key factors of dataset diversity. Second,
we present evaluated datasets and settings. Next, we explore these
factors independently and compare them. Finally, we give some discussions.

\subsection{Factors of Dataset Diversity}

The dataset diversity can be reflected in two aspects including instance
density and category diversity. On one hand, dense instances can provide
each concept with lots of variations in pose, scale, illumination,
distortion, background, etc. Thus the FSIR model can learn more accurate
knowledge from dense instances. On the other hand, diverse categories
can provide plenty of concepts with larger visual difference. Thus
the FSIR model can learn more comprehensive knowledge from diverse
categories.

\subsection{Datasets and Settings}

As ImageNet \cite{russakovsky2015imagenet} is a well-known large
dataset which has been widely used for both visual recognition and
FSIR (e.g., MiniImageNet \cite{vinyals2016matching}), we construct
different subsets which contain dense instances and diverse categories
for FSIR. We use the ImageNet (ILSVRC2012) dataset with 1000 categories
and 1.28 million images. Each category corresponds to one synset in
WordNet \cite{miller1995wordnet}. We find its parent synsets recursively
until reaching the node of entity which is the root of the WordNet,
according to their hierarchical semantic relations. In this manner,
we can obtain the entire tree structure of the 1000 categories, which
are divided into different branches with different relevance.

\begin{figure}
\centering{}\includegraphics[scale=0.6]{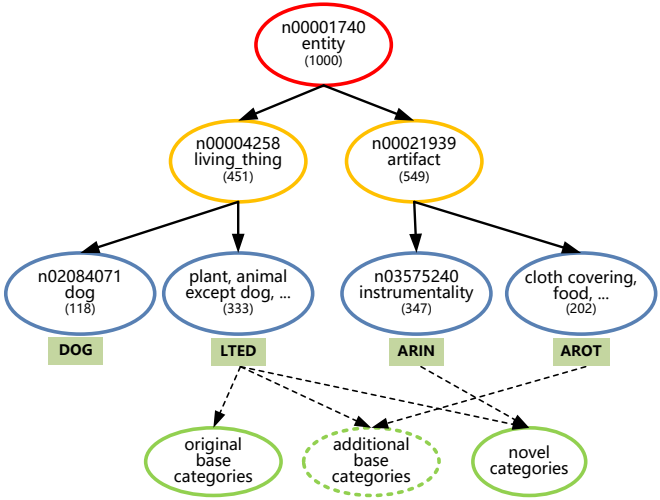} \caption{The hierarchical display of the 1000 categories in ImageNet. The integer
in the parenthesis indicates the number of the categories that the
branch contains. The number in the beginning of the node is the corresponding
WordNet ID. The LTED, ARIN and AROT are three branches, which are
sampled to form original (additional) base and novel categories.}
\label{fig:The-hierarchical-display} 
\end{figure}

The branches of the 1000 categories are illustrated in Fig. \ref{fig:The-hierarchical-display}.
All categories belong to the entity branch, as the entity node is
the root of the tree structure. And they are divided into two branches
according to whether they are man-made or nature. More precisely,
as shown in Fig. \ref{fig:The-hierarchical-display}, the first one
is living thing branch which contains 451 categories, and the second
one is artifact branch which contains 549 categories. The living thing
branch is further divided into the dog branch (which contains various
dogs, denoted as DOG) and the LTED (which is the short for Living
Thing Except Dog) branch which is the rest of the living thing branch
except the categories in the dog branch. Meanwhile, the artifact branch
is divided into the ARIN (which is short for ARtifact INstrument)
branch which is the super class of instrumentality and the AROT (which
is short for ARtifact Other Thing) branch which is the rest of the
artifact branch.

To qualitatively measure the relevance between different categories,
we utilize the approaches used in \cite{li2019multifaceted,yosinski2014transferable}
to estimate their relevance, according to the tree structure of ImageNet.
The categories in the same branch are more relevant than the ones
in other different branches. For example, as the categories in the
LTED branch contain animals such as cat, sheep, kangaroo, while categories
in the ARIN branch contain many traffic instruments, the base networks
trained on some categories belonging to LTED have features which will
help classify some other categories in LTED branch. So the categories
in the LTED branch are more relevant to each other compared to the
categorizes in the other branches.

Meanwhile, measuring the relevance between different datasets quantitatively
is complementary to the qualitative analysis, and it can distinguish
different datasets more clearly and accurately. So we utilize Word
Mover's Distance (WMD)\ \cite{wmd-2015,chen2020graph} and Shortest
Path Length (SPL)\ \cite{kutuzov2019making,miller1995wordnet,raada1989development}
to measure their relevance quantitatively. Specifically, a small distance
means a great relevance. 
\begin{itemize}
\item WMD: the WMD distance originally measures the dissimilarity between
two text documents as the minimum amount of distance that the embedded
words of one document need to travel to reach the embedded words of
another document. Recently, it has been proposed for distributional
metric matching and applied to cross-domain alignment. In the same
spirit of the approach used in \cite{chen2020graph}, we use WMD to
measure the relevance between two sets of entities. Let $\mathrm{\mathbb{D}}_{x}$
and $\mathrm{\mathbb{D}}_{y}$ represent two sets of categories
(i.e., entities) from two different datasets. As each category in
the data corresponds to one synset in WordNet, we utilize the corresponding
text to represent each category. In this way, $\mathrm{\mathbb{D}}_{x}$
and $\mathrm{\mathbb{D}}_{y}$ are denoted as $\tilde{X}=\{\tilde{x}_{i}\}_{i=1}^{n}$
and $\tilde{Y}=\{\tilde{y}_{i}\}_{i=1}^{m}$ respectively, where $n$
and $m$ are the number of categories in each dataset and $\tilde{x}_{i}$
and $\tilde{y}_{i}$ are the corresponding words. The distance between
$\mathrm{\mathrm{\mathbb{{D}}}}_{x}$ and $\mathrm{\mathrm{\mathbb{{D}}}}_{y}$
is obtained by computing the dissimilarity between $\tilde{X}$ and
$\tilde{Y}$ with the WMD metric, denoted as $\mathcal{D}_{wmd}(\mathrm{\mathrm{\mathbb{{D}}}}_{x},\mathrm{\mathrm{\mathbb{{D}}}}_{y})=WMD(\tilde{X},\tilde{Y})$.
Here we utilize the word2vec\ \cite{mikolov2013distributed} embedding
to represent each word in $\tilde{X}$ and $\tilde{Y}$. 
\item SPL: the main idea of path-based measures is that the similarity between
two concepts is a function of the length of the path linking the concepts
and the position of the concepts in the taxonomy. The SPL distance
which is a variant of the distance methods of\ \cite{raada1989development,bulskov2002measuring}
has been widely used to measure the semantic similarities of different
concepts in WordNet\ \cite{miller1995wordnet}. As the SPL based
on WordNet has shown its talents and attracted great concern, we also
use SPL to measure the relevance between two sets of concepts. Depending
on the structure of WordNet, generally the result obtained from hypernym
relation is regarded as the similarity between concepts. Let us define
the length of the shortest path from synset $c_{i}$ to synset $c_{j}$
in WordNet as $len(c_{i},c_{j})$, then $len(c_{i},c_{j})$ is counted
as the actual path length between $c_{i}$ and $c_{j}$. For example,
if $c_{i}$ and $c_{j}$ are not the same node, but $c_{i}$ is the
parent of $c_{j}$, we assign the semantic length between them to
1, i.e., $len(c_{i},c_{j})=1$. In this way, for one category $c_{xi}$
in the dataset $\mathrm{\mathrm{\mathbb{{D}}}}_{x}$, its SPL distance
to the dataset $\mathrm{\mathrm{\mathbb{{D}}}}_{y}$ is denoted as
$d(c_{xi},\mathrm{\mathrm{\mathbb{{D}}}}_{y})=\sideset{\frac{1}{M}}{_{j=1}^{M}}\sum len(c_{xi},c_{yj})$,
where $M$ is the number of categories in $\mathrm{\mathrm{\mathbb{{D}}}}_{y}$.
Finally, the SPL distance between $\mathrm{\mathrm{\mathbb{{D}}}}_{x}$
and $\mathrm{\mathrm{\mathbb{{D}}}}_{y}$ is $\mathcal{D}_{spl}(\mathrm{\mathrm{\mathbb{{D}}}}_{x},\mathrm{\mathrm{\mathbb{{D}}}}_{y})=\sideset{\frac{1}{N}}{_{i=1}^{N}}\sum d(c_{xi},\mathrm{\mathrm{\mathbb{{D}}}}_{y})$,
where $N$ is the number of categories in $\mathrm{\mathrm{\mathbb{{D}}}}_{x}$. 
\end{itemize}
\begin{table}
\caption{\label{tab:Some-SOTA-FSL-1}Distances between different datasets which
are obtained with the metric of WMD.}
 
\begin{centering}
\begin{tabular}{|c|c|c|c|c|}
\hline 
 & DOG  & LTED  & ARIN  & AROT\tabularnewline
\hline 
DOG  & 0  & 3.1981  & 3.5829  & 3.5438\tabularnewline
\hline 
LTED  & 3.1981  & 0  & 3.5102  & 3.4782\tabularnewline
\hline 
ARIN  & 3.5829  & 3.5102  & 0  & 3.3787\tabularnewline
\hline 
AROT  & 3.5438  & 3.4782  & 3.3787  & 0\tabularnewline
\hline 
\end{tabular}
\par\end{centering}
 
\end{table}

\begin{table}
\caption{\label{tab:Some-SOTA-FSL-1-1}Distances between different datasets
which are obtained by the metric of SPL according to the tree structure
of WordNet.}
 
\centering{}%
\begin{tabular}{|c|c|c|c|c|}
\hline 
 & DOG  & LTED  & ARIN  & AROT\tabularnewline
\hline 
DOG  & 0  & 13.39  & 19.60  & 18.97\tabularnewline
\hline 
LTED  & 13.39  & 0  & 16.12  & 15.40\tabularnewline
\hline 
ARIN  & 19.60  & 16.12  & 0  & 10.80\tabularnewline
\hline 
AROT  & 18.97  & 15.40  & 10.80  & 0\tabularnewline
\hline 
\end{tabular}
\end{table}

For the WMD distance, the results are shown in Table\ \ref{tab:Some-SOTA-FSL-1}.
We utilize the approach\ \cite{wmd-2015} under the setting where
the words are represented with the word2vec\ \cite{mikolov2013distributed}
embedding vectors. The WMD distance between two categories is a specific
case where each set only contains one entity. For example, the WMD
distance between the entity "basset" (i.e., n02088238) and the entity
"beagle" (i.e., n02088364) which both belong to the DOG branch is
3.0119. The WMD distance between the entity "basset" and the entity
"flattop" (i.e., n02687172 ) is 4.0327, which is much bigger as
the entity "flattop" belongs to the ARIN branch. It is illustrated
that "basset" is close to "beagle" while "basset" is far away
from "flattop". As these categories are special entities selected
from different divided branches of ImageNet, the distance between
two sets of categories will be within these ranges. So it is can be
concluded that the WMD distance between two sets roughly varies from
3 to 5. For a holistic view, the WMD distances among the DOG, LTED,
ARIN, and AROT branches are shown in Table\ \ref{tab:Some-SOTA-FSL-1}.
The results demonstrate that the categories in the some branch have
bigger relevance. As the results show, the WMD distance between DOG
and LTED, denoted as $\mathcal{D}_{wmd}(DOG,LTED)$, is 3.1981. However,
$\mathcal{D}_{wmd}(DOG,ARIN)$ and $\mathcal{D}_{wmd}(DOG,AROT)$
are 3.5829 and 3.5438 respectively, which are much bigger compared
with $\mathcal{D}_{wmd}(DOG,LTED)$. Because DOG and LTED are corresponded
to the living thing branch, they are more relevant with each other,
compared to ARIN and AROT which are covered by the artifact branch.
Meanwhile, $\mathcal{D}_{wmd}(ARIN,AROT)$ is 3.3787, while $\mathcal{D}_{wmd}(ARIN,LTED)$
is 3.5102. Because $\mathcal{D}_{wmd}(ARIN,LTED)$ is much bigger
than $\mathcal{D}_{wmd}(ARIN,AROT)$, it is demonstrated that ARIN
is more relevant with AROT, rather than LTED.

The results based on the SPL metric are shown in Table\ \ref{tab:Some-SOTA-FSL-1-1}.
The SPL distance between the entity "basset" (i.e., n02088238) and
the entity "beagle" (i.e., n02088364) 
is 2. Note that the entity "hound" (i.e., n02087551) is the parent
node of both "basset" and "beagle". The SPL distance between the
entity "basset" and the entity "flattop" (i.e., n02687172) is
22, 
as the entity "flattop" belongs to the ARIN branch. From these cases,
it also can be concluded that the SPL distance between two sets roughly
varies from 2 to 22. The SPL distances among the DOG, LTED, ARIN,
and AROT branches are shown in Table\ \ref{tab:Some-SOTA-FSL-1-1}.
The SPL distance between DOG and LTED, denoted as $\mathcal{D}_{spl}(DOG,LTED)$
is 13.39. However, $\mathcal{D}_{spl}(DOG,ARIN)$ and $\mathcal{D}_{spl}(DOG,AROT)$
are 16.12 and 15.40 respectively, which are much bigger than $\mathcal{D}_{spl}(DOG,LTED)$.
Meanwhile, $\mathcal{D}_{spl}(ARIN,AROT)$ is 10.80, while $\mathcal{D}_{spl}(ARIN,LTED)$
and $\mathcal{D}_{spl}(AROT,LTED)$ are 16.12 and 15.40 respectively.
Both of them are much bigger than $\mathcal{D}_{spl}(ARIN,AROT)$.
The trend is the same with the one under the measure of WMD, as shown
in Table\ \ref{tab:Some-SOTA-FSL-1}.

All the results under two metrics illustrate that the datasets covered
by the same branch have smaller distances, demonstrating that the
categories in the some branch have bigger relevance. Furthermore,
the quantitative results are consistent with the qualitative results.
WMD measures the relevances among different datasets or branches based
on the semantic embedding vectors of the composed entities, while
SPL measures the relevance of them depending on how close of the entities
are in the taxonomy. They depict the relevance among different datasets
complementarily, and they offer and establish baseline distributional
metrics for comparing sets of concepts in computer vision tasks.

Base and novel categories are sampled from the LTED, ARIN, and AROT
branches to explore the dataset diversity for FSIR. In this section,
we do not use the DOG branch since this branch lacks of diversities
of images. The few-shot learning model would suffer from handling
a sequence of training tasks originated from different distributions
if the novel categories are irrelevant to the base categories. In
the following sub-sections, we carry out various groups of experiments,
based on different base and novel categories. A group of experiments
includes original or with additional base categories and novel categories,
as illustrated in Fig \ref{fig:The-hierarchical-display}. Each group
of experiments is conducted 5 times with eight few-shot learning methods
(e.g., PN \cite{snell2017prototypical}, RN \cite{sung2018learning},
DSN \cite{simon2020adaptive}, DEMD \cite{zhang2020deepemd}, MAML
\cite{Finn2017Model}, adaCNN \cite{Munkhdalai2018Rapid}, Proto-MAML
\cite{meta-2020}, MetaOpt \cite{lee2019meta}) to obtain stable and
reliable performance. The eight methods use a 4 convolutional layers
as the meta-learner (backbone) with the different number of filters
per layer. We adopt the architectures of corresponding methods without
modification for our experiments. Without loss of generality, we analyze
the factors of dataset diversity on 1-shot models. To evaluate each
model, 400 test tasks are randomly sampled from 20 novel categories.
And each test task has 5 classes, each of which has 1 image in the
support set and 15 images in the target set. These test settings have
been widely used in the few-shot evaluation \cite{vinyals2016matching,snell2017prototypical,Finn2017Model,Munkhdalai2018Rapid,Munkhdalai2017Meta,sung2018learning},
and the results are reported with mean accuracy.

\subsection{Instance Density}

\begin{figure}
\centering{}\includegraphics[scale=0.27]{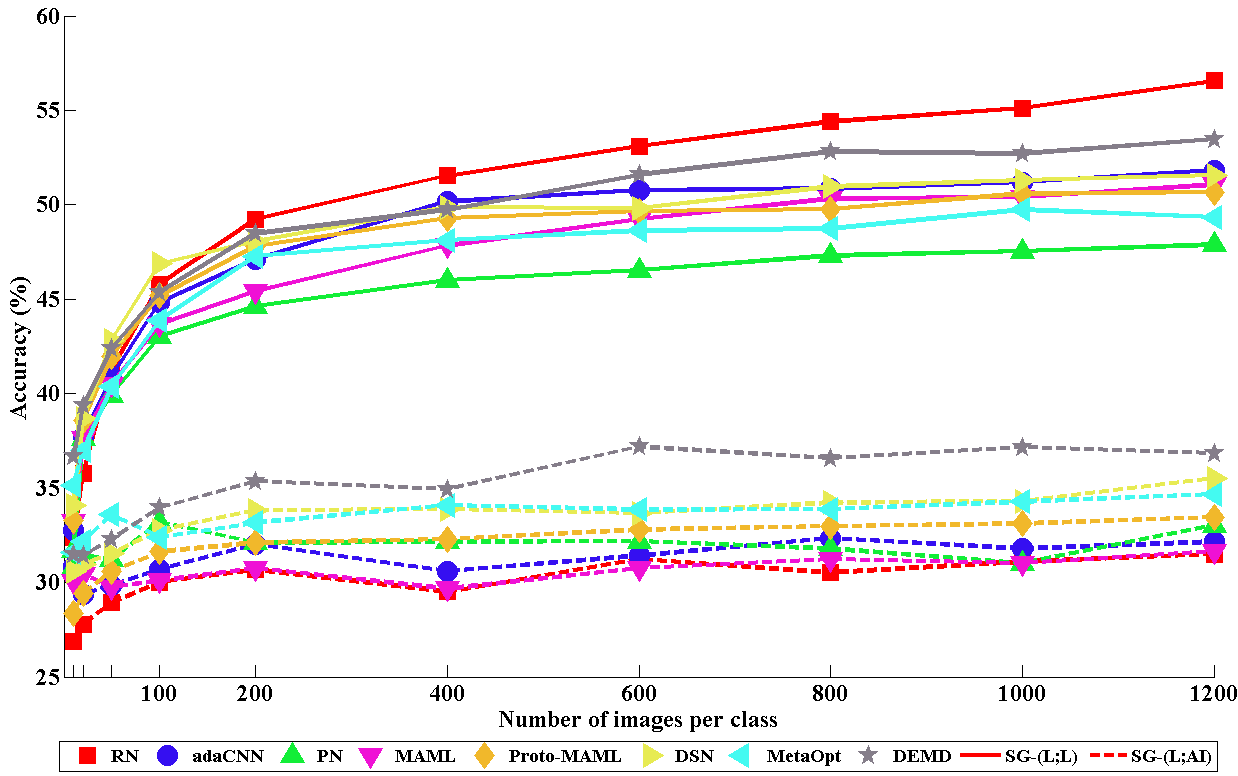} \caption{\textcolor{black}{5-way 1-shot accuracy of SG-(L;L) and SG-(L;AI).}}
\label{fig:Experimental-results-of} 
\end{figure}

This subsection studies the problem of whether increasing the instance
density can bring better performance on test tasks. Different from
the study of the impact of the shot number \cite{cao2020theoretical},
this subsection aims to investigate the impact of performance as the
number of samples per base category increases. We conduct two groups
of experiments with the samples growth (SG) per base category, and
each of them uses 64 base categories with the number of samples per
category ranging from 10 to 1,200. One group uses base and novel categories
sampled from LTED branch, denoted as SG-(L;L). Another group uses
the same base categories, but it employs novel categories sampled
from ARIN branch, denoted as SG-(L;AI). Each novel category in SG-(L;L)
has more relevant base categories than that in SG-(L;AI). In other
words, SG-(L;L) uses relevant base categories while SG-(L;AI) employs
irrelevant ones.

Fig. \ref{fig:Experimental-results-of} displays the performance of
two groups. It can be observed that: i) The performance of SG-(L;L)
exceeds SG-(L;AI) by a wide margin with the same methods, especially
as the number of samples per base category increases. ii) In SG-(L;L),
more instances lead to better performance, and the performance improvement
is fast when the number of samples per base category ranges from 10
to 200, while it gets slow when starting from 200. iii) In SG-(L;AI),
the performance is improved at the beginning of the number of instances
increasing (from 10 to 100), after which the performance is not significantly
improved.

The following suggestions can be obtained. 
\begin{itemize}
\item It is very important to use plenty of instances from relevant base
categories to train the FSIR model. 
\item If the relevant base categories are not available, there is no need
to use too many instances of each category. 
\end{itemize}

\subsection{Category Diversity}

This subsection studies whether increasing the category diversity
can bring better performance on test tasks. Since the number of base
categories is changed, we divide base categories into original and
additional ones. We set up two groups of experiments with category
growth (CG), and each of them uses 64 original base categories and
varying additional ones. One group employs original, additional base
categories and novel categories all sampled from LTED branch, denoted
as CG-(L,L;L). Another group exploits the same original base and novel
categories, but it uses additional base categories sampled from AROT
branch, denoted as CG-(L,AO;L). It is obvious that CG-(L,L;L) uses
relevant additional base categories while CG-(L,AO;L) employs irrelevant
ones. The number of samples per original or additional base category
is 600. The samples of original base categories are fixed in the same
group, and the number of additional base categories ranges from 0
to 64.

\begin{figure}
\centering{}\includegraphics[scale=0.3]{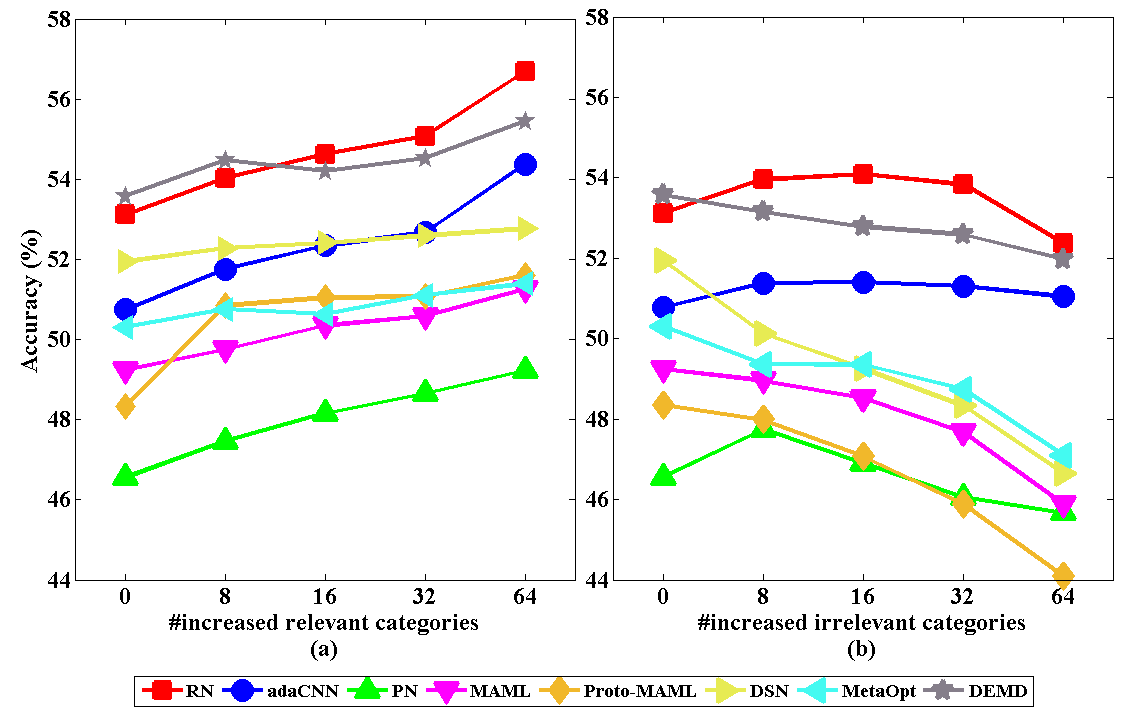} \caption{\textcolor{black}{5-way 1-shot accuracy of CG-(L,L;L) (a) and CG-(L,AO;L)
(b).}}
\label{fig:Experimental-results-of-test LI} 
\end{figure}

\begin{table}
\caption{\label{tab:Sampled-categories-of}Sampled categories of different
groups.}

\noindent \centering{}%
\begin{tabular}{|l|c|c|c|}
\hline 
\multirow{1}{*}{Groups} & \multicolumn{2}{c|}{Base categories} & \multirow{1}{*}{Novel categories}\tabularnewline
 & Original  & Additional  & \tabularnewline
\hline 
\multicolumn{1}{|l|}{SG-(L;L)} & LTED  & -  & LTED\tabularnewline
\multicolumn{1}{|l|}{SG-(L;AI)} & LTED  & -  & ARIN\tabularnewline
\hline 
\multicolumn{1}{|l|}{CG-(L,L;L)} & LTED  & LTED  & LTED\tabularnewline
\multicolumn{1}{|l|}{CG-(L,AO;L)} & LTED  & AROT  & LTED\tabularnewline
\multicolumn{1}{|l|}{CG-(L,L;AI)} & LTED  & LTED  & ARIN\tabularnewline
\multicolumn{1}{|l|}{CG-(L,AO;AI)} & LTED  & AROT  & ARIN\tabularnewline
\hline 
\multicolumn{1}{|l|}{CGS-(L,L;L)} & LTED  & LTED  & LTED\tabularnewline
\multicolumn{1}{|l|}{CGS-(L,L;AI)} & LTED  & LTED  & ARIN\tabularnewline
\multicolumn{1}{|l|}{CGS-(L,AO;L)} & LTED  & AROT  & LTED\tabularnewline
\multicolumn{1}{|l|}{CGS-(L,AO;AI)} & LTED  & AROT  & ARIN\tabularnewline
\hline 
\end{tabular}
\end{table}

Experimental results of CG-(L,L;L) and CG-(L,AO;L) on eight methods
are illustrated in Fig. \ref{fig:Experimental-results-of-test LI}
(a) and (b), respectively. It can be observed that: i) When additional
base categories are relevant to novel categories, more additional
categories lead to better performance (see Fig. \ref{fig:Experimental-results-of-test LI}
(a)). ii) When additional base categories and novel categories are
irrelevant, the performance may be improved when the number of categories
increases at the beginning, after which the performance drops (see
Fig. \ref{fig:Experimental-results-of-test LI} (b)).

On the other hand, we set another two groups of experiments denoted
as CG-(L,L;AI) and CG-(L,AO;AI). Different from the above two groups,
novel categories of them are sampled from ARIN branch, as illustrated
in Table \ref{tab:Sampled-categories-of}. Obviously, CG-(L,L;AI)
uses irrelevant additional base categories while CG-(L,AO;AI) employs
relevant ones.

\begin{figure}
\noindent \begin{centering}
\includegraphics[scale=0.3]{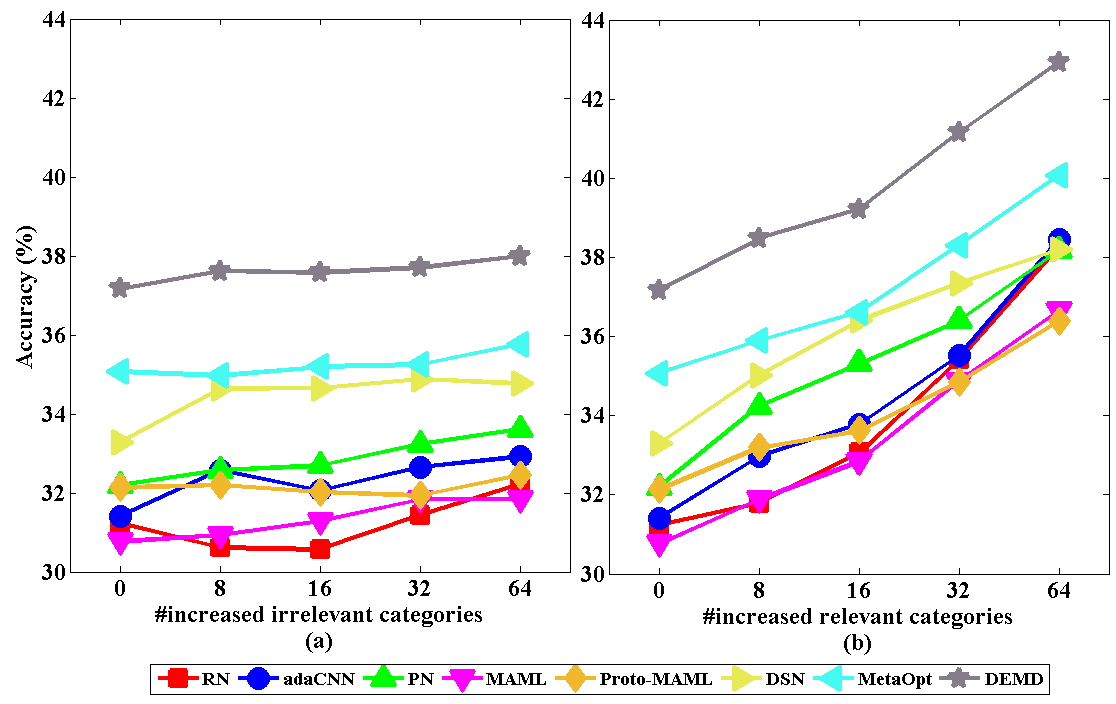} 
\par\end{centering}
\caption{\label{fig:Experimental-results-of-test Ari}\textcolor{black}{5-way
1-shot accuracy of CG-(L,L;AI) (a) and CG-(L,AO;AI) (b).}}
\end{figure}

Experimental results of CG-(L,L;AI) and CG-(L,AO;AI) on eight methods
are illustrated in Fig. \ref{fig:Experimental-results-of-test Ari}
(a) and (b), respectively. It can be observed that: i) The performance
of CG-(L,AO;AI) is obviously superior to CG-(L,L;AI) as the number
of additional base categories increases. The main reason is that CG-(L,AO;AI)
uses relevant additional base categories while CG-(L,L;AI) uses irrelevant
ones. ii) The performance of two groups is improved as the number
of additional base categories increases. In addition, comparing CG-(L,L;L)
and CG-(L,AO;AI) (or CG-(L,AO;L) and CG-(L,L;AI)), relevant original
base categories provide better initial performance than irrelevant
ones. 

The following suggestions can be obtained. 
\begin{itemize}
\item It is very important to use plenty of relevant base categories to
learn the FSIR model. 
\item If original base categories are not relevant to novel categories,
we can use some additional base categories without constraints. 
\end{itemize}

\subsection{Instance Density v.s. Category Diversity}

From the above experiments, the FSIR performance would be improved
by increasing relevant instances or relevant categories of base categories.
To further explore which factor is more effective to boost the performance,
we make comparisons by fixing the number of total samples while varying
the number of categories and the number of samples in each category.
In this case, the number of samples in each category will be decreased
to guarantee the same total number of samples, as the number of base
categories increases. Since the relevance of base and novel categories
would affect the performance, we set up two groups of experiments
with the same relevance of original and additional base categories,
and each of them includes 32 original base categories and varying
additional ones. One group employs categories all sampled from LTED
branch, denoted as CGS-(L,L;L) (Category Growth under the Same total
samples). Another group exploits the same original and additional
base categories, but it uses novel categories sampled from ARIN branch,
denoted as CGS-(L,L;AI). Thus CGS-(L,L;L) uses relevant base categories
while CGS-(L,L;AI) employs irrelevant ones. Original base categories
are fixed in the same group of experiments, and the number of additional
base categories ranges from 0 to 64. The total number of samples of
base categories is 38,400 (equal to the number of total training samples
in MiniImageNet), where each base category contains equal number of
samples. Since the total number of samples of base categories is fixed,
each original base category would have less samples with the growth
of the number of additional base categories. For example, each original
base category contains 1,200 samples without additional base categories
(the total number is 1200{*}32=38,400), and the number of samples
of each original base category would reduce to 800 when 16 additional
base categories with 800 samples per category are used (the total
number is 800{*}(32+16)=38,400).

\begin{figure}
\centering{}\includegraphics[scale=0.3]{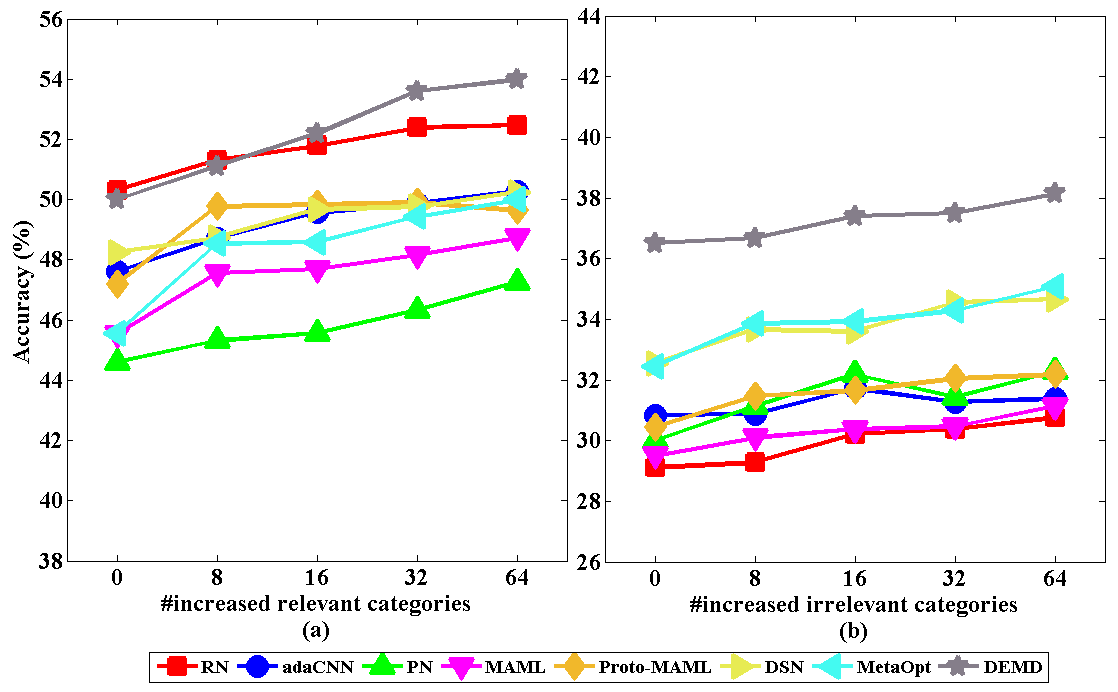} \caption{\textcolor{black}{5-way 1-shot accuracy of CGS-(L,L;L) (a) and CGS-(L,L;AI)
(b).}}
\label{fig:Experimental-results-of-varying training12} 
\end{figure}

Experimental results of CGS-(L,L;L) and CGS-(L,L;AI) on eight methods
are illustrated in Fig. \ref{fig:Experimental-results-of-varying training12}
(a) and (b), respectively. It can be observed that increasing base
categories is more effective than increasing their samples to boost
the FSIR performance, when original and additional base categories
are relevant. That is to say, it is better for the FSIR model to use
more relevant base categories than more samples per relevant base
category (obtained from Fig. \ref{fig:Experimental-results-of-varying training12}
(a)), and it is better for the FSIR model to use more irrelevant base
categories than more samples per irrelevant base category (obtained
from Fig. \ref{fig:Experimental-results-of-varying training12} (b)).
This phenomenon explains that additional base categories provide more
bonus for the FSIR model than learned categories with additional samples,
since the model has already learned accurate knowledge from hundreds
of samples per base category.

Moreover, we set another two groups of experiments with irrelevance
of original and additional base categories. The two groups are denoted
as CGS-(L,AO;L) and CGS-(L,AO;AI), which are different from the above
two groups since they use additional base categories sampled from
AROT branch, as illustrated in Table \ref{tab:Sampled-categories-of}.
Obviously, CGS-(L,AO;L) uses irrelevant additional base categories
while CGS-(L,AO;AI) employs relevant additional ones, and CGS-(L,AO;L)
uses relevant original base categories while CGS-(L,AO;AI) employs
irrelevant original ones.

\begin{figure}
\centering{}\includegraphics[scale=0.3]{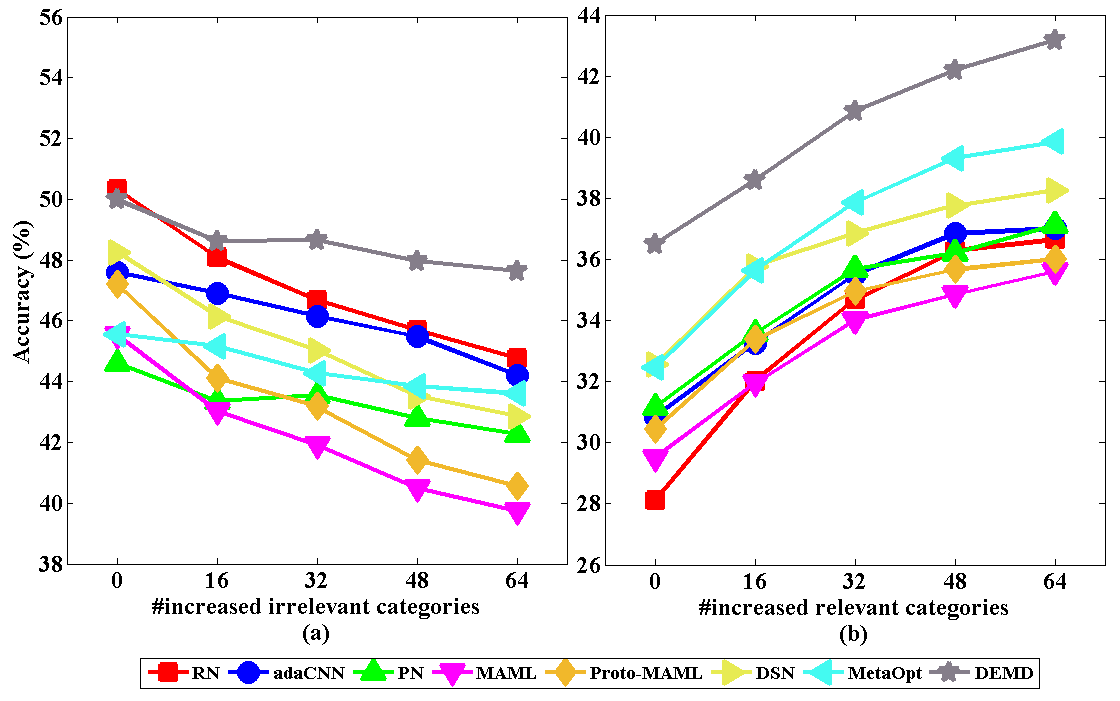} \caption{\textcolor{black}{5-way 1-shot accuracy of CGS-(L,AO;L) (a) and CGS-(L,AO;AI)
(b).}}
\label{fig:Experimental-results-of-varying training34} 
\end{figure}

Experimental results of CGS-(L,AO;L) and CGS-(L,AO;AI) on eight methods
are illustrated in Fig. \ref{fig:Experimental-results-of-varying training34}
(a) and (b), respectively. From Fig. \ref{fig:Experimental-results-of-varying training34}
(a), it is better for the FSIR model to use more samples per relevant
base category than more irrelevant base categories. From Fig. \ref{fig:Experimental-results-of-varying training34}
(b), it is better for the FSIR model to use more relevant base categories
than more samples per irrelevant base category. These two observations
can explain that relevance is a more important factor than more base
categories or more sample per base category.

The following suggestions can be obtained. 
\begin{itemize}
\item It is better for the FSIR model to use more base categories than more
samples per category, when original and additional base categories
are relevant. 
\item Relevance is a more important factor than transformations in sample
forms (i.e., more categories or more sample per category). 
\end{itemize}
According to the above two points, we can further infer that the FSIR
performance goes down in order of more relevant base categories, more
samples per relevant base category, more irrelevant base categories,
and more samples per irrelevant base category.

\subsection{Tremendous Number of Categories}

When the number of total base samples is fixed,whether increasing
the number of base categories will improve performance? To explore
this problem, we conduct experiments on a dataset of a larger number
of categories. This dataset contains 10,020 categories, sampled from
ImageNet \cite{russakovsky2015imagenet}. In this case, we set the
total number of samples as 20,000, and the number of base categories
ranges from 100 to 10,000. Thus each category contains samples ranging
from 200 to 2. For each setting (indexed with the number of base categories),
the evaluations are also conducted 5 times with eight few-shot learning
methods. For the evaluation of each model,{} we use the same architecture
of few-shot learning methods, the same number of test tasks or novel
categories, and evaluation index as the above experiments.

\begin{figure}
\begin{centering}
\includegraphics[scale=0.35]{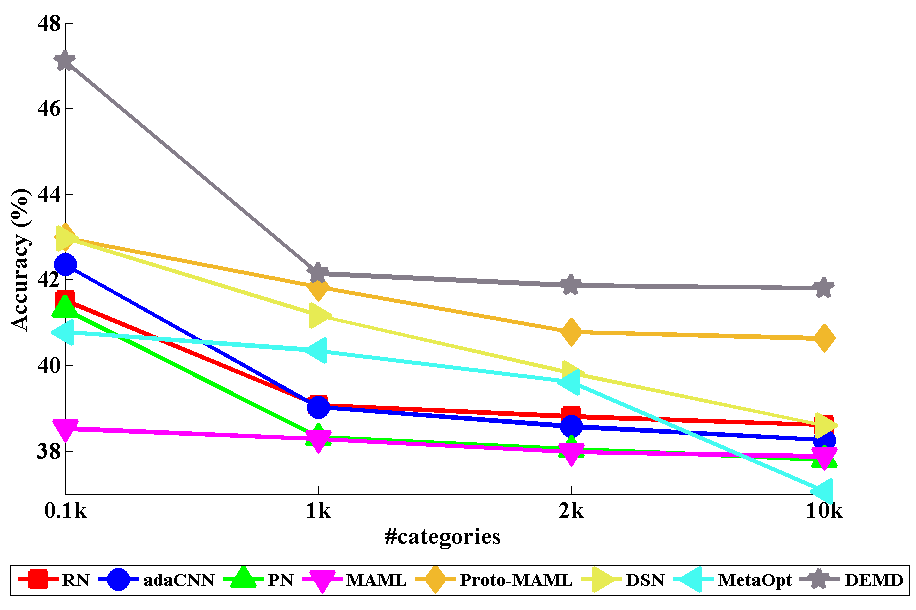} 
\par\end{centering}
\caption{\label{fig:Experimental-results-on a}\textcolor{black}{Experimental
results on a large number of categories with a fixed amount of total
samples.} }
\end{figure}

Experimental results on eight methods are illustrated in Fig. \ref{fig:Experimental-results-on a}.
With the fixed number of total samples, as the number of base categories
increases, the FSIR performance decreases. When the number of per
base category is 1,000, namely each base category contains 20 samples,
more samples per base category become more important than more base
categories. In this case, additional samples per base category provide
more bonus for the FSIR model than additional base categories, since
the FSIR model can not learn accurate knowledge well from a small
amount of samples per category. 

\subsection{Discussion}

The main factors about base categories affecting the performance of
FSIR are presented as follows: 
\begin{itemize}
\item The more categories are not always the better. Relevance is the key
factor, and increasing relevant categories can benefit FSIR, otherwise,
including irrelevant categories may not be helpful. 
\item Data diversity is also an important factor to FSIR, and large diversity
in categories or density in instances can result in better results,
enhancing the generalization capability of the FSIR model.
\item When enough instances are in each category, category diversity is
more sensitive than instance density, which may coarsely broaden the
diversity of the whole dataset. 
\end{itemize}

\section{\label{sec:Evaluation-of-Dataset}Evaluation of Dataset Structure}

In early studies, FSIR methods attempt to recognize the alphabet images
\cite{Lake2015Human} or the object images \cite{fei2006one} with
simple background. Afterwards, FSIR methods focus on more realistic
images, such as general object images \cite{vinyals2016matching,snell2017prototypical,hariharan2017low,sung2018learning},
fine-grained object images \cite{vinyals2016matching,wei2019piecewise},
scene images \cite{Dixit2017AGA,schwartz2018delta}. However, different
type of images have intrinsic properties and data structures, which
may lead to significant performance differences. In this section,
we study the dataset structure through the view of FSIR. First, we
present various factors of dataset structures with quantitative representations.
Second, we introduce several datasets under few-shot settings. Finally,
we make analysis on different datasets from the dataset structure
and few-shot learning methods.

\subsection{\label{subsec:Factors-of-dataset}Factors of Dataset Structure}

\begin{figure}
\noindent \begin{centering}
\includegraphics[width=0.9\columnwidth]{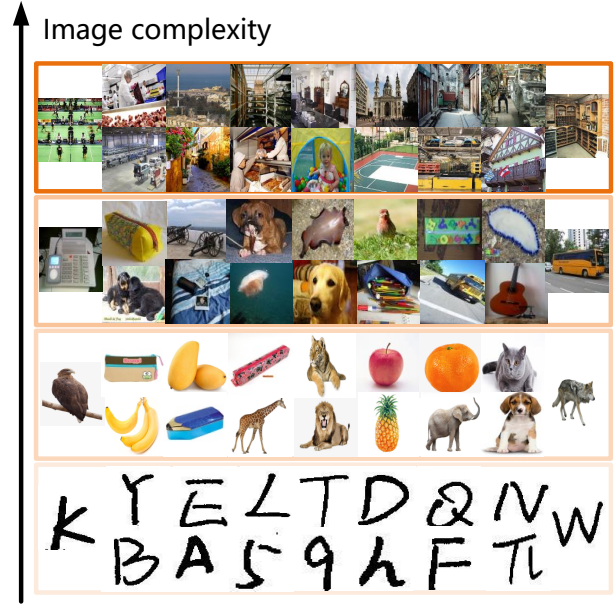} 
\par\end{centering}
\caption{\label{fig:The-rank-of}Example illustrations of different levels
of image complexity.}
\end{figure}

Dataset structure can be reflected with image complexity, intra-concept
visual consistency and inter-concept visual similarity. Image complexity
can depict visual contents of original images. If original images
include complex background information, it could be difficult to accurately
identify their concepts. As illustrated in Fig.\ \ref{fig:The-rank-of},
the apple image is more complex than the letter ``L'' image, and
the scene image with cluster background is more complex than the apple
image. Both intra-concept visual consistency and inter-concept visual
similarity can depict semantic gaps between low-level visual features
(i.e., computational representations of images from hand-crafted algorithms)
and high-level image concepts (i.e., semantic interpretations of images
from human beings) \cite{fan2012quantitative}. Intra-concept visual
consistency describes the aggregations of single concept in the visual
feature space, and a big intra-concept visual consistency may result
in low semantic gaps. In contrast, inter-concept visual similarity
describes correlations between different concepts in the visual feature
space, and a small inter-concept visual similarity may result in low
semantic gaps. When considering intra-concept visual consistency and
inter-concept visual similarity simultaneously, we obtain that a larger
difference value between them means a smaller semantic gap.

We quantify image complexity with the following ways: i) Complex images
often have complex structures, and structural attributes can be preserved
by edges. We use percentage of edge points $P_{EP}$ as a metric,
defined as: 
\begin{equation}
P_{EP}=\frac{\#(\mathrm{edge\,pixels})}{\#(\mathrm{image\,pixels})}
\end{equation}
where $\#(\mathrm{image\,pixels})$ and $\#(\mathrm{edge\,pixels})$
are the number of pixels and edges in the image, respectively. Edge
pixels can be detected with the Canny edge detection method \cite{canny1986computational}.
ii) We utilize the 2D entropy $E_{2D}$ \cite{larkin2016reflections}
to measure the underlying spatial structure and pixel co-occurrence.
Given an $N\times M$ pixel image $f(n,m)$, $E_{2D}$ is defined
as: 
\begin{equation}
\begin{split}E_{2D}=-\sum_{i=I_{min}}^{I_{max}}\sum_{j=J_{min}}^{J_{max}}p_{i,j}log_{2}p_{i,j}\end{split}
\end{equation}
where $p_{i,j}=\frac{1}{N\times M}\sum_{n=1}^{N}\sum_{m=1}^{M}\delta_{i,f_{x}(n,m)}\delta_{j,f_{y}(n,m)}$,
$\delta$ is the Kronecker delta formulation, $f_{x}$ and $f_{y}$
are the two derivative components of the gradient field, $I$ and
$J$ record all gradient values in two gradient directions.

Intra-concept visual consistency and inter-concept visual similarity
are quantified according to \cite{fan2012quantitative}. Intra-concept
visual consistency of category $C_{l}$ is defined as: $c_{ina}(C_{l})=\frac{2}{|C_{l}|*(|C_{l}|-1)}\sum_{i=1}^{|C_{l}}\sum_{j=i+1}^{|C_{l}|}k(x_{l}^{i},x_{l}^{j})$,
where $|C_{l}|$ is the number of images in $C_{l}$, $x_{l}^{i}$
and $x_{l}^{j}$ are the feature representations of images in $C_{l}$.
$k(x_{l}^{i},x_{l}^{j})=\nicefrac{c}{\sqrt{|\frac{x_{l}^{i}-u}{\sigma}-\frac{x_{l}^{j}-u}{\sigma}|^{2}+c}}$,
where $c$ is a scalar, $k(x_{l}^{i},x_{k}^{j})$ is inversely proportional
to the Euclidean distance of $\frac{x_{l}^{i}-u}{\sigma}$ and $\frac{x_{l}^{j}-u}{\sigma}$
($\mu$ and $\sigma$ are the mean and standard deviation of all $x$,
$\frac{x_{l}^{i}-u}{\sigma}$ can be regarded as Z-score normalization
of all $x$), and can be used as the similarity of them. The overall
intra-concept visual consistency $C_{ina}$ of all categories in $D$
is defined as: 
\begin{equation}
C_{ina}=\frac{1}{|D|}\sum_{l=1}^{|D|}c_{ina}(C_{l})
\end{equation}
where $|D|$ is the number of categories in $D$. Inter-concept visual
similarity between $C_{l}$ and $C_{k}$ is defined as: $s_{inr}(C_{l},C_{k})=\frac{1}{|C_{l}||C_{k}|}\sum_{i=1}^{|C_{l}|}\sum_{j=1}^{|C_{k}|}k(x_{l}^{i},x_{k}^{j})$,
where $|C_{l}|$ and $|C_{k}|$ are the number of images of $C_{l}$
and $C_{k}$, respectively, $x_{l}^{i}$ and $x_{k}^{j}$ are the
feature representations of images in $C_{l}$ and $C_{k}$. The overall
inter-concept visual similarity $S_{inr}$ of each two categories
in $D$ is defined as: 
\begin{equation}
S_{inr}=\frac{2}{|D|(|D|-1)}\sum_{l=1}^{|D|}\sum_{k=l+1}^{|D|}s_{inr}(C_{l},C_{k})
\end{equation}
We use three types of classical features (i.e., GIST: used in \cite{Oliva2001Modeling},
HOG: Histograms of Oriented Gradients \cite{dalal2005histograms},
LBP: Local Binary Patterns \cite{ojala2002multiresolution}.) to calculate
$C_{ina}$ and $S_{inr}$, and set $c=1.0$. We believe that classical
descriptors can somewhat convey the internal structure of datasets
from multiple aspects. In addition to the classical descriptors, we
also include convolutional neural network (CNN) and bag-of-word (BoW)
features for comparison, which may be biased to the training data.
In our case CNN model is pretrained on ImageNet, codebook of BoW is
learned within each dataset. As illustrated in Table \ref{tab:Datasets-complexity-of},
the trends of the statistical results on different datasets are almost
the same in most cases, making the analysis be more comprehensive.
It can be observed that in all cases (using different types of features
in different datasets), the intra-concept similarity is larger than
inter-concept similarity. HOG and LBP descriptors focus more on local
details, which are sensitive to the content of dataset, resulting
in larger deviation among different datasets. In contrast, GIST, BoW,
and CNN focus more on global information, where more consistent results
are obtained in different datasets.

\begin{figure*}
\centering{}\includegraphics[scale=0.78]{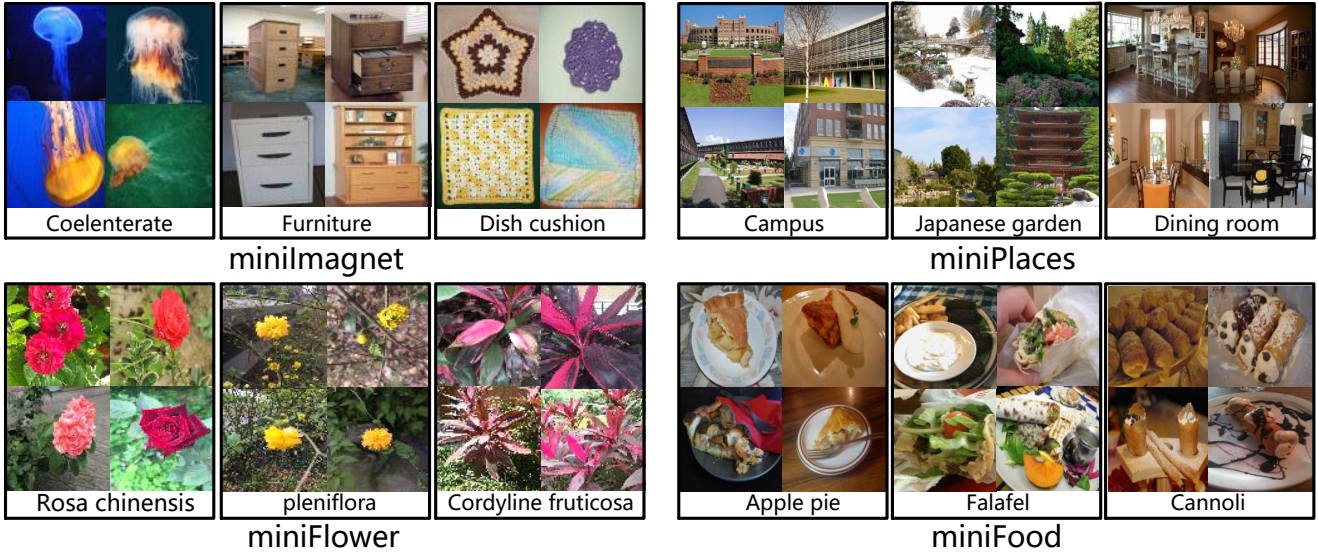} \caption{Some image examples in different datasets.}
\label{fig:Some-Images-in} 
\end{figure*}

\subsection{\label{subsec:Evaluated-datasets}Datasets and Settings}

We construct several datasets with different structures, which range
from simple character images to the images with one or more objects.
In addition to the above images, facing vertical fields, some fine-grained
datasets (i.e., food and flower datasets) are also used in our evaluations.
To balance the diversity of different datasets, we refer to MiniImagenet,
and organize the datasets with the same number of categories and the
same number of samplers per category as MiniImagenet. These datasets
are collected as follows. i)\textbf{ MiniCharacter} is handwritten
character dataset, which is generated and annotated by 15 volunteers.
MiniCharacter includes various characters such as English letters,
numbers, mathematical symbols. The total number of categories is 84,
and each of category has 100 images. The 64 and 20 categories are
used as base and novel categories, respectively. ii)\textbf{ MiniImagenet}
\cite{vinyals2016matching} uses 64 and 20 categories as base and
novel categories, respectively, where each category contains 600 images.
iii) \textbf{MiniPlaces} is a subset of Places365 \cite{zhou2018places}.
It uses 64 and 20 categories as base and novel categories, respectively,
where each category contains 600 images. iv)\textbf{ MiniFlower} is
sampled from flower dataset \cite{favcFlowers} provided from 2018
FGCVxFlower Classification Challenge. It uses 64, 20 categories as
base, novel categories, respectively, where each category contains
600 images. v)\textbf{ MiniFood} is sampled from Food101 \cite{bossard2014food}.
It uses 64 and 20 categories as base and novel categories, respectively,
where each category contains 600 images.  Experiments are conducted
with eight few-shot learning methods (i.e., PN, RN, DSN, DEMD, MAML,
adaCNN, Proto-MAML, MetaOpt). Evaluating each model, we use the same
number of test tasks, evaluation index and the same architecture of
few-shot learning methods ( denoted as CNN-4), as Sect. \ref{sec:Evaluation-of-Diversity}.
Besides, we also use a deeper architecture ResNet-12, which has been
widely adopted in recent works \cite{Munkhdalai2018Rapid,oreshkin2018tadam,simon2020adaptive,zhang2020deepemd,lee2019meta}. 

\begin{table}
\centering{}\caption{The image complexity in different datasets.}
\label{tab:IC} %
\begin{tabular}{l|c|c}
\hline 
\multirow{1}{*}{Datasets} & $P_{EP}$  & $E_{2D}$\tabularnewline
\hline 
MiniCharacter  & 0.048  & 2.190\tabularnewline
MiniImagenet  & 0.225  & 7.045\tabularnewline
MiniPlaces  & 0.254  & 7.411\tabularnewline
MiniFlower  & 0.325  & 8.012\tabularnewline
MiniFood  & 0.259  & 7.407\tabularnewline
\hline 
\end{tabular}
\end{table}

\subsection{Analysis on Dataset Structure}

\begin{table*}
\centering{}\caption{5-way 1-shot accuracy (\%) on different datasets with CNN-4. }
\label{tab:Experimental-results-on} %
\begin{tabular}{l|c|c|c|c|c|c|c|c}
\hline 
\multirow{2}{*}{Datasets} & \multicolumn{4}{c|}{Metric-based methods} & \multicolumn{4}{c}{Meta-learning methods}\tabularnewline
\cline{2-9} \cline{3-9} \cline{4-9} \cline{5-9} \cline{6-9} \cline{7-9} \cline{8-9} \cline{9-9} 
 & PN  & RN  & DSN  & DEMD  & MAML  & adaCNN  & Proto-MAML  & MetaOpt\tabularnewline
\hline 
MiniCharacter  & 86.17  & 91.83  & 87.21  & 88.43  & 85.45  & 83.91  & 88.00  & 88.80\tabularnewline
MiniImagenet  & 49.42  & 51.38  & 51.03  & 52.46  & 48.70  & 48.26  & 48.41  & 51.67\tabularnewline
MiniPlaces  & 49.00  & 50.62  & 51.89  & 53.96  & 48.49  & 50.05  & 50.44  & 55.74\tabularnewline
MiniFlower  & 48.73  & 50.73  & 51.07  & 55.24  & 49.32  & 48.56  & 50.60  & 51.95\tabularnewline
MiniFood  & 41.41  & 45.68  & 44.86  & 50.70  & 43.55  & 44.94  & 43.63  & 45.73\tabularnewline
\hline 
\end{tabular}
\end{table*}

\begin{table*}
\caption{\label{tab:Experimental-results_resnet12}5-way 1-shot accuracy (\%)
on different datasets with ResNet-12.}

\begin{centering}
\begin{tabular}{l|c|c|c|c|c|c|c|c}
\hline 
\multirow{2}{*}{Datasets} & \multicolumn{4}{c|}{Metric-based methods} & \multicolumn{4}{c}{Meta-learning methods}\tabularnewline
\cline{2-9} \cline{3-9} \cline{4-9} \cline{5-9} \cline{6-9} \cline{7-9} \cline{8-9} \cline{9-9} 
 & PN  & RN  & DSN  & DEMD  & MAML  & adaCNN  & Proto-MAML  & MetaOpt\tabularnewline
\hline 
MiniCharacter  & 92.87  & 92.15  & 92.70  & 93.49  & 85.65  & 89.26  & 94.24  & 93.10\tabularnewline
MiniImagenet  & 58.47  & 53.08  & 58.67  & 60.46  & 54.90  & 57.13  & 57.37  & 59.73\tabularnewline
MiniPlaces  & 60.55  & 58.01  & 61.30  & 60.47  & 58.25  & 56.35  & 60.4  & 63.12\tabularnewline
MiniFlower  & 55.63  & 55.36  & 55.26  & 61.62  & 54.15  & 58.10  & 55.13  & 58.53\tabularnewline
MiniFood  & 51.35  & 49.29  & 51.58  & 54.76  & 49.44  & 52.15  & 50.63  & 52.30\tabularnewline
\hline 
\end{tabular}
\par\end{centering}
 
\end{table*}

Our goal is to analyze the performance of different dataset from their
structures. Table \ref{tab:IC} illustrates the image complexity of
different datasets. which is measured with $P_{EP}$ and $E_{2D}$.
Table \ref{tab:Experimental-results-on} and Table \ref{tab:Experimental-results_resnet12}
show the results of different methods on different datasets. It can
be observed that the image complexity in MiniCharacter is the lowest.
Each character image contains only one character with clear background.
Table \ref{tab:Experimental-results-on} also demonstrates that the
performance on MiniCharacter is significantly higher than the one
on the other datasets. From this special case of MiniCharacter, we
can find that image complexity plays an important role in FSIR, where
lower complexity often leads to higher performance. However, character
images are very different with most real-world images, such as object,
scene and food images. In the rest evaluations, we mainly focus on
comparisons and discussions of the other four datasets.

As illustrated in Table \ref{tab:IC}, MiniFlower is the highest in
image complexity. However, the FSIR performance on MiniFlower is not
the lowest (see Table \ref{tab:Experimental-results-on} and Table
\ref{tab:Experimental-results_resnet12}), which is higher than MiniFood
and comparable with MiniImagenet. This may be explained from the following
reasons: i) MiniFlower is a special kind of fine-grained image dataset,
where images in the same category are very similar. As shown in Fig.
\ref{fig:Some-Images-in}, although image complexity is very high
in the aspects of clustered edges and detailed component information,
the visual texture and structural patterns of images in the same category
are still very similar. MiniFlower has the highest intra-concept visual
consistency $C_{ina}$, as illustrated in Table \ref{tab:Datasets-complexity-of}.
This may present that the task of flower recognition is relatively
easier, even for the case of FSIR. ii) As categories in MiniFlower
all belong to flower and they are biological relatives, it is intuitive
that visual patterns of different flower categories are also similar
to some extent, as shown in Fig. \ref{fig:Some-Images-in}. Thus base
categories and novel categories in miniFlower present some relevance,
which enables learned FSIR models to better recognize novel categories.

Except for the very simple dataset MiniCharacter, most methods obtain
the highest performance on MiniPlaces (see Table \ref{tab:Experimental-results-on}
and Table \ref{tab:Experimental-results_resnet12}). MiniFlower is
more complex than MiniPlaces (see Table \ref{tab:IC}), which seems
to be the reason that most methods obtain lower performances on MiniFlower
than that on MiniPlaces. MiniPlaces has comparable image complexity
to the MiniFood, however, the average gains of all methods on MiniPlaces
(over MiniFood) are close to 6.8\% in Table \ref{tab:Experimental-results-on}
and 7.8\% in Table \ref{tab:Experimental-results_resnet12}. In alternative
to the absolute complexity, such difference in accuracy may be caused
by the difference value between intra-concept visual consistency and
inter-concept visual similarity with CNN features in Table \ref{tab:Datasets-complexity-of},
where MiniPlaces dataset has a larger difference value. Similarly,
MiniImagenet has a smaller difference value than MiniPlaces, thus,
most methods still obtain higher performances on MiniPlaces compared
to MiniImagenet. 

\begin{table*}
\centering{}\caption{\label{tab:Datasets-complexity-of}Intra-concept visual consistency
and inter-concept visual similarity in different datasets.}
\begin{tabular}{l|ccc|cc|ccc|cc}
\hline 
\multirow{2}{*}{Datasets} & \multicolumn{5}{c|}{$C_{ina}$ (\%)} & \multicolumn{5}{c}{$S_{inr}$ (\%)}\tabularnewline
 & GIST  & BoW  & CNN  & HOG  & LBP  & GIST  & BoW  & CNN  & HOG  & LBP\tabularnewline
\hline 
MiniImagenet  & 0.45  & 2.40  & 5.11  & 0.25  & 11.24  & 0.37(0.08)  & 2.23(0.17)  & 4.90(0.21)  & 0.20(0.05)  & 7.64(3.60)\tabularnewline
MiniPlaces  & 0.44  & 2.40  & 4.96  & 0.27  & 4.22  & 0.36(0.08)  & 2.23(0.17)  & 4.73(0.23)  & 0.20(0.07)  & 3.63(0.59)\tabularnewline
MiniFlower  & 0.55  & 2.38  & 5.29  & 0.44  & 18.91  & 0.46(0.09)  & 2.22(0.16)  & 5.11(0.18)  & 0.37(0.07)  & 17.74(1.17)\tabularnewline
MiniFood  & 0.45  & 2.40  & 5.16  & 0.31  & 11.67  & 0.39(0.06)  & 2.23(0.07)  & 4.97(0.19)  & 0.25(0.06)  & 11.24(0.43)\tabularnewline
\hline 
\end{tabular}
\end{table*}

The FSIR performance on MiniImagenet is higher than MiniFood and comparable
with MiniFlower. The main reason is that image complexity in MininImagenet
is lower than the one in the two datasets. However, the advantage
of performance on MiniImagenet is faint or not obvious when considering
its image complexity is obviously lower than that of MiniFlower. This
phenomenon can be attributed to less relevance of base categories
and novel categories in MiniImagenet, which is reflected in the following
two aspects. i) Category labels in MiniImageNet are sampled from a
larger semantic space which brings greater diversities of semantic
concepts, compared with MiniFlower. This leads to less probabilities
of sampling relevant base and novel categories in MiniImagenet, when
the number of base categories or novel categories are the same in
all datasets (i.e., all datasets include 64 base categories and 20
novel categories). ii) Images in MiniImagenet focus on single object,
and different categories differ greatly in visual content, in contrast,
different categories in MiniFlower are similar, as shown in Fig. \ref{fig:Some-Images-in}.
Thus base categories and novel categories in MiniImagenet present
less relevance than that in MiniFlower.

The FSIR performance from all methods on MiniFood is the worst. The
reasons can be explained as follows. i) As shown in Table \ref{tab:IC},
the image complexity of MiniFood is much higher than the one of MiniImagenet
and comparable with MiniPlaces. Besides, compared with the two datasets,
the inter-concept visual similarity $S_{inr}$ of MiniFood is higher
(see Table \ref{tab:Datasets-complexity-of}). Therefore, the performance
on MiniFood is worse than the one on MiniImagenet and MiniPlaces.
ii) Although MiniFlower and MiniFood both belong to fine-grained datasets,
there are fixed semantic patterns in MiniFlower. For example, the
flower consists of some semantic parts, such as petals and calyx.
However, such semantic patterns do not exist in minFood \cite{min2019a},
and minFood thus has a lower intra-concept visual consistency $C_{ina}$
than MiniFlower (see Table \ref{tab:Datasets-complexity-of}). Therefore,
the FSIR performance on MiniFood is worse than the one on MiniFlower.

\begin{figure*}
\begin{centering}
\begin{tabular}{cc}
\includegraphics[scale=0.36]{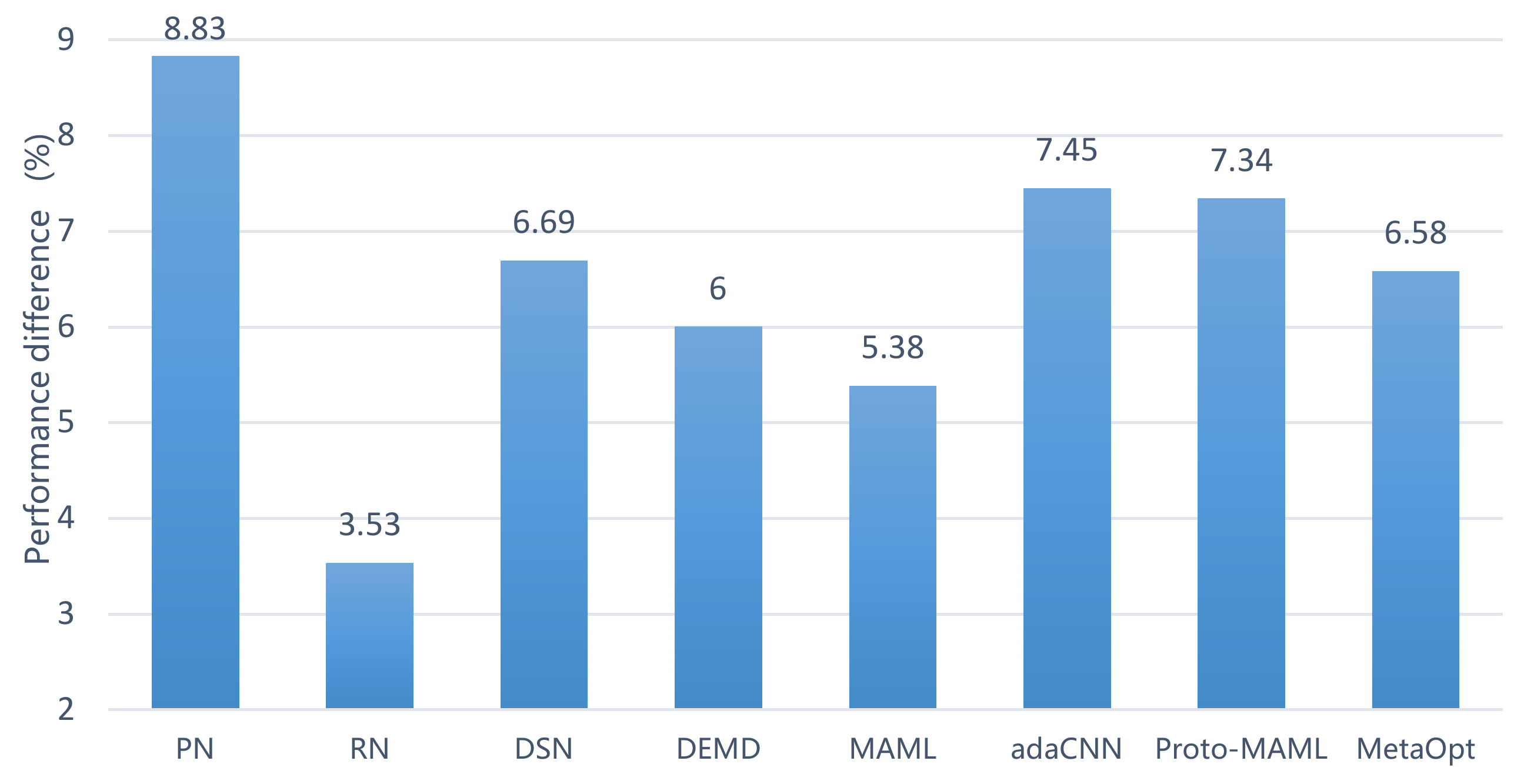}  & \includegraphics[scale=0.42]{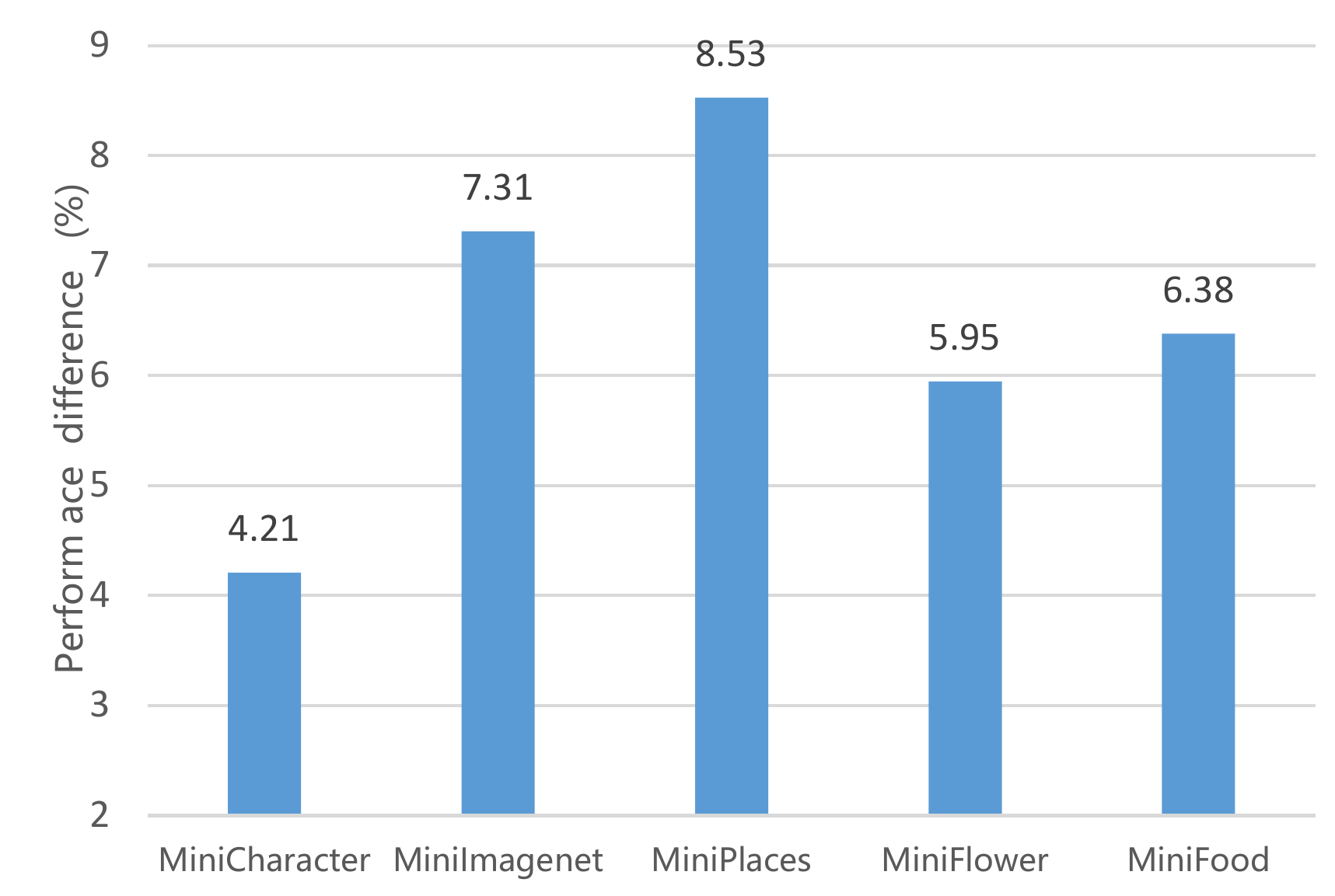} \tabularnewline
(a)  & (b) \tabularnewline
\end{tabular}
\par\end{centering}
\centering{}\caption{\label{fig:average_gain} (a): the average gain from CNN-4 to ResNet-12
over all datasets for different methods. (b): the average gain from
CNN-4 to ResNet-12 over all methods on different datasets.}
\end{figure*}

\subsection{Analysis on Few-shot Learning Methods}

 As shown in Fig. \ref{fig:average_gain} (a), for all methods, with
a deeper backbone ResNet-12, PN obtains the largest performance gain
averaged on the five datasets (8.83\%). The probable reason is that
a deeper network makes PN learn better prototypical representation
compared with shallow networks. In contrast, the averaged gain from
all datasets is the lowest (3.53\%) for RN. RN can obtain powerful
features from the deep backbone ResNet-12. In this case, a simple
metric is enough for effective few-shot learning. However, RN adopts
a more complex metric implemented by convolutional transformation,
leading to the limited generalization ability. Both PN and RN belong
to metric-based methods, while among different meta-learning methods,
MAML obtains the smallest averaged gain from the backbone of CNN-4
to ReNet-12. Compared to adaCNN and MetaOpt, MAML needs to learn the
same initialization, from which different tasks can achieve effective
adaptation with gradient descent. However, compared to the shallow
backbone (CNN-4), it's more difficult to obtain better initialization
on the deeper backbone (ResNet-12), due to the larger parameter space
for initialization. In contrast, Proto-MAML initializes the classifier
with the class prototype, achieving a good initialization when using
the backbone ResNet-12. Thus the gain of Proto-MAML is much higher
than MAML, and comparable to adaCNN and MetaOpt. As shown in Fig.
\ref{fig:average_gain} (b), All methods obtain the performance gain
from one shallow backbone (CNN-4) to a deeper backbone (ResNet-12).
Particularly, the average gain of all methods on MiniCharacter is
the smallest (4.21\%), which is intuitive, as images in MiniCharcter
is very simple. The average gain on MiniPlaces is the biggest (8.53\%).
The reason is that images in MiniPlaces usually contain multiple objects,
using deeper architectures with more convolutional and pooling operations
can progressively generate more abstract feature representations for
this kind of images. 

In metric-base methods, DSN and DEMD outperform classical metric-based
methods PN and RN, benefiting from the more effective metric learning
methods. In particular, DSN learns task-adaptive different subspace
metrics, and DEMD uses the Earth Mover's Distance to capture detailed
differences of each two images in local regions. And compared with
other meta-learning methods, such as adaCNN, MAML and Proto-MAML,
MetaOpt obtains better accuracy on most datasets (except for miniCharacter).
In particular, MetaOpt can share common parameters (in backbone models)
for different tasks, allowing to focus on the optimizing of high-level
classifiers, leading to fast adaptation with only few shots in training.
Among metric-based/meta-learning methods, DEMD/MetaOpt obtains better
results on most datasets. When comparing with best meta-learning (MetaOpt)
and metric-based (DEMD) methods, the former works better on datasets
with more objects in larger difference value between intra-concept
consistency and inter-concept similarity, such as MiniPlaces, while
the latter works better on more fine-grained datasets, such as MiniFlower
and MiniFood. DEMD and MetaOpt obtain comparable results on more simple
datasets with smaller complexity (in Table \ref{tab:Datasets-complexity-of}),
such as MiniCharacter and MiniImageNet. Since miniPlaces dataset usually
consists of object co-occurrences between different categories, sharing
the common parameters and focusing on high-level discrimination allow
MetaOpt to obtain better performance on miniPlaces. Since different
categories in fine-grained datasets usually contain similar prototypes,
DEMD can focus more on detailed regions, leading to better performance
on more fine-grained datasets.

One interesting method is Proto-MAML, a meta-learning method like
MAML, whose classifier is initialized with the class prototype from
metric-based method PN. Thus, the comparison between ProtoMAML, MAML
and PN is worth discussing. The results of Proto-MAML on MiniImagenet
is in line with the reported results in \cite{meta-2020}, where the
accuracy of Proto-MAML is slightly worse than PN. However, compared
with MAML and PN, Proto-MAML obtains better performance on three datasets
(i.e., MiniCharacter, MiniPlaces, and MiniFood) with CNN-4, and on
MiniCharacter with ResNet-12. With a shallow backbone (CNN-4), Proto-MAML
works better than PN and MAML on most datasets, demonstrating that
a suitable initialization (with PN in Proto-MAML) is also necessary
to MAML. Otherwise, the trained model may fall into some suboptimal
states due to fast convergence in the shallow architecture. In contrast,
with a deeper backbone (ResNet-12), PN works slightly better than
Proto-MAML on most datasets, showing that PN can be better optimized
with a deeper architecture. Furthermore, when using ResNet-12, MAML
obtains a lower performance than PN on all datasets. That's probably
because it is difficult for MAML to learn a good initialization in
a large parameter space. Compared to MAML, Proto-MAML can be optimized
with better initialization, allowing it to obtain more gains on most
datasets. 

\subsection{\label{subsec:Discussion}Discussion}

Based on the above analysis, we can obtain: 
\begin{itemize}
\item A dataset with a low image complexity presents better performance.
Besides, intra-concept visual consistency and inter-concept visual
similarity also can influence performance differences.
\item Different methods have diverse performance on different types of datasets,
which is relevant to two important dimensions, namely dataset structures
and the method ability. According to our experimental analysis, the
best metric-based method and the best meta-learning one obtain comparable
performance on different datasets. A better metric function is important
for metric-based methods while better adaptive strategies are vital
for meta-learning methods. Combining better metric function learning
with better adaptive strategies can be explored in FSIR.
\item It's more effective to design FSIR architectures according to the
characteristics of data distribution. For instance, DEMD focuses more
on detailed regions, resulting in better performance on fine-grained
datasets. MetaOpt shares common parameters and focuses on optimizing
category sensitive classifiers, obtaining best accuracy on miniPlaces
with object co-occurrences between different scenes. 
\end{itemize}

\section{\label{sec:Perspectives-and-future}Perspectives and future directions}

According to existing FSIR researches and our observations, we give
some prospective analysis on future works from the following aspects.

\textbf{Transferable FSIR.} It is common to utilize prior representations
obtained in models which are trained with large examples \cite{dvornik2020selecting}
for new tasks. Our work has demonstrated that the relevance between
images in base and novel categories, and the data structures of the
images have much influence on the FSIR performance. The results provide
criterion for selecting or integrating features for FSIR. Furthermore,
there are situations that images in base categories and novel categories
have different data distributions \cite{chen2019closer,new-bench-2019}.
Sometimes, images in novel categories are even coupled with unlabeled
data \cite{phoo2020self}. It is important for few-shot learning methods
to eliminate the domain gap between base and novel categories, thus
these methods can obtain representations with better generalization
ability. The analysis of the dataset bias somewhat reveals the shared
information among different examples. These results also provide useful
guidance for designing transductive FSIR models to further improve
FSIR performance. Moreover, in real world FSIR scenarios, how to explore
knowledge from available images to recognize images in some specific
target domains is in high demand. For example, recognizing fine-grained
novel categories \cite{tang2020revisiting,zhu2020multi} uses generic
base categories instead of fine-grained ones, considering the data
from general images is easier to access. Our work has investigated
the transferable capability impacted by the dataset bias, introducing
insightful observations to effectively obtain adapted representations
for FSIR. 

\textbf{Incremental learning of few shot learning}.  In many applications,
it is often desirable to have the flexibility of learning novel concepts,
with limited data and without re-training on the full training set,
namely Incremental Few-Shot Learning (IFSL) \cite{ren2019incremental,yoon2020xtarnet,tao2020few,liu2020incremental}.
In this task, the novel classes with only a few labeled examples for
each class should be considered based on the trained model for a set
of base classes. Similar to existing few-shot learning methods, IFSL
should also consider the recognition performance on unseen novel classes.
Therefore, some observations from our work can still provide guidance
in designing IFSL. For example, if the unseen novel classes are very
relevant to based categories, IFSLs can obtain better recognition
performance. On the other hand, different from existing few-shot learning
methods, IFSL should also consider the recognition performance on
base classes. For that, incremental leaning is incorporated into the
few-shot learning framework. The performance differences on different
datasets from dataset structures and different few-shot learning methods
are worth exploration. For example, what types of few-shot learning
methods lead to better performance of IFSL when combined with incremental
learning? Our experimental design and analysis can provide the reference
for supporting such exploration. For example, integrating metric-based
methods and meta-learning ones based on their different learning mechanisms
is worthy of further study.

\textbf{FSIR on data of non-uniform distributions}. Data in non-uniform
distribution, such as data in long-tailed distribution (LTD) and out-of-
distribution (OOD), is typically challenging to the machine learning
models. The vast majority of existing FSIR works assume training data
of each base category is even. However, this is a highly restrictive
setting, LTD is the most common in real-world scenarios, where some
categories have plenty of samples while more categories have only
a few ones. One goal of few-shot learning is to address the problem
in the recognition with LTD data. Particularly in our work, we demonstrate
that larger data (without increasing categories) and diverse category
distribution may enhance the generative ability of the model. In addition
to long-tailed data distribution, a more challenging task is to deal
with the OOD data, where test data is not in the same distribution
as training data. How to use this kind of data of non-uniform distributions
for FSIR is worth exploring. In our case, we demonstrate that different
datasets have various complexity, intra-concept consistency and inter-concept
similarity, thus, different datasets are OOD to each other. Mixing
those OOD data may damage transferability of the FSIR model, thus,
how to filter those noise from the data with mixed distributions is
worth researching. Moreover, it's more practical and challenging to
use unlabeled OOD data for data augmentation, thus, few-shot learning
with semi-supervised training is also a significant research direction
in future. 

\textbf{FSIR with extra knowledge}. Since visual data contains complex
structural information, it is difficult to directly extract effective
knowledge, especially for the generic knowledge which can be used
to establish some connections on novel categories. However, this generic
knowledge can be relatively easily obtained from some prior knowledge
such as attributes \cite{Dixit2017AGA,zhu2020attribute}, word embeddings
(learned from large-scale text corpora) \cite{wang2017multi,peng2019few,chen2020knowledge},
relationships of entities \cite{peng2019few,chen2020knowledge,li2019large},
knowledge graph \cite{peng2019few,chen2020knowledge} etc. This kind
of information are complementary for FSIR, and the problem of dataset
bias can be alleviated to some extent by applying techniques of information
fusion or knowledge transfer. Extra knowledge is useful both for the
cases of category diversity or dataset from different domains. As
human can easily correlate various kinds of extra knowledge to recognize
a new object, we believe FSIR with extra knowledge is a potentially
important research topic to improve performance of existing few shot
benchmarks as well as for practical applications.

\section{\label{sec:Conclusions}Conclusions}

Few-shot image recognition (FSIR) is an important research problem
in the machine learning and computer vision community. In this paper,
we study FSIR with dataset bias systematically. First, we investigate
impact of transferable capabilities learned from distributions of
base categories. We introduce instance density and category diversity
to depict distributions of base categories, and relevance to measure
relationships between base categories and novel categories. Experimental
results on different sub-datasets of ImageNet demonstrate the relevance,
instance density and category diversity can depict the transferable
bias from base categories. Second, we investigate differences in performance
on different datasets from characteristics of dataset structures and
different few-shot learning methods. We introduce image complexity,
intra-concept visual consistency, and inter-concept visual similarity
to quantify characteristics of dataset structures. From comprehensive
experimental evaluations on five datasets with eight few-shot learning
methods, some insightful observations are obtained from the perspective
of both dataset structures and few-shot learning methods. We hope
these observations are helpful to guide future FSIR research.



{\small{}{} \bibliographystyle{IEEEtran}
\bibliography{ms}
 }{\small\par}

\end{document}